\documentclass[lettersize,journal]{IEEEtran}
\usepackage{amsmath,amsfonts}
\usepackage{algorithmic}
\usepackage{algorithm}

\usepackage{array}
\usepackage[caption=false,font=normalsize,labelfont=sf,textfont=sf]{subfig}
\usepackage{textcomp}
\usepackage{stfloats}
\usepackage{url}
\usepackage{booktabs}
\usepackage{verbatim}
\usepackage{graphicx}
\usepackage{cite}
\usepackage{caption}
\usepackage{color}
\usepackage{multirow}
\usepackage[colorlinks,bookmarksopen,bookmarksnumbered,citecolor=blue, linkcolor=blue, urlcolor=black]{hyperref}

\usepackage{makecell}

\begin{document}

\title{TSceneJAL: Joint Active Learning of Traffic Scenes for 3D Object Detection}
\author{Chenyang Lei$^1$, Weiyuan Peng$^1$, Guang Zhou$^{4}$, Meiying Zhang$^1$, Qi Hao$^{1}$, Chunlin Ji$^{3}$, Chengzhong Xu$^{2}$
\thanks{Chenyang Lei, Weiyuan Peng and Guang Zhou are co-first authors of the article; Corresponding author: Meiying Zhang (e-mail: zhangmy@sustech.edu.cn), and Qi Hao (e-mail: Haoq@sustech.edu.cn)}
\thanks{$^1$ Research Institute of Trustworthy Autonomous Systems, and Department of Computer Science and Engineering, Southern University of Science and Technology (SUSTech), China. $^2$ State Key Lab of Internet of Things for Smart City, University of Macau, China. $^3$ Kuang-Chi Institute of Advanced Technology, China; $^4$ Shenzhen Deeproute.ai Co.,Ltd, China.}
}

\markboth{Journal of \LaTeX\ Class Files,~Vol.~14, No.~8, August~2021}%
{Shell \MakeLowercase{\textit{et al.}}: A Sample Article Using IEEEtran.cls for IEEE Journals}


\maketitle

\begin{abstract}
Most autonomous driving (AD) datasets incur substantial costs for collection and labeling, inevitably yielding a plethora of low-quality and redundant data instances, thereby compromising performance and efficiency. Many applications in AD systems necessitate high-quality training datasets using both existing datasets and newly collected data. In this paper, we propose  a traffic scene joint active learning (TSceneJAL) framework that can efficiently sample the balanced, diverse, and complex traffic scenes from both labeled and unlabeled data. The novelty of this framework is threefold:  1) a scene sampling scheme based on a category entropy, to identify scenes containing multiple object classes, thus mitigating class imbalance for the active learner; 2) a similarity sampling scheme, estimated through the directed graph representation and a marginalize kernel algorithm, to pick sparse and diverse scenes; 3) an uncertainty sampling scheme, predicted by a mixture density network, to select instances with the most unclear or complex regression outcomes for the learner. Finally, the integration of these three schemes in a joint selection strategy yields an optimal and valuable subdataset. {Experiments on the KITTI, Lyft, nuScenes and SUScape datasets demonstrate that our approach outperforms existing state-of-the-art methods on 3D object detection tasks with up to 12\% improvements.}


\end{abstract}


\section{Introduction}

\IEEEPARstart{D}{\lowercase{{ata}-driven}} 3D object detection for AD systems relies heavily on extensive training data. However, the process of collecting and annotating AD scenes is both time-consuming and costly. Moreover, high-dimensional datasets like KITTI\cite{geiger2012we}, Lyft\cite{kes2019lyft}, and SUScape\cite{suscape} introduce significant amounts of irrelevant, redundant, and noisy information, escalating computational overhead and impairing perception performance. 
Therefore, some semi-supervised/weakly supervised methods \cite{zhao2020sess,wang20213dioumatch,li2023dds3d,meng2020weakly,xu2022back} have been proposed, involving training with limited precise/coarse labels. But, these approaches typically overlook the evaluation of unlabeled data and are constrained to specific scenarios. To streamline the collection and assessment of the data instances for AD tasks, many active learning (AL) strategies \cite{sener2017active,feng2019deep, ash2019deep, luo2023exploring} have been introduced. 
Most AL frameworks consist of four essential components: labeled/unlabeled data, predictor, sampler, and oracle. Initially, the predictor is trained using labeled dataset to generate pesudo-labels for unlabeled data. Subsequently, the sampler leverages evaluation metrics \cite{liu2022survey} to identify the most informative data from the unlabeled dataset. These selected instances can then be labeled by the oracle, utilizing annotation tools like \cite{li2020sustech,wu2023efficient}.   
The crux of such frameworks lies in the formulation of effective evaluation metrics that guide the sampler in selecting samples to enhance the predictor's performance. In general, a fully functional AL approach to 3D object detection should be able to address the following three key issues, as depicted in Fig. \ref{fig:intro}:
\begin{figure}  
    \centering     
    \includegraphics[width=0.9\linewidth]{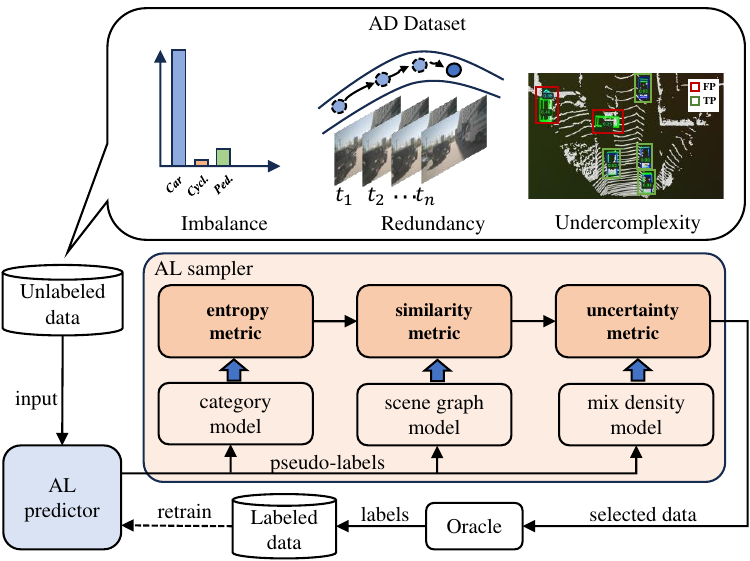}     
    \caption{A brief illustration of the proposed AL framework. The AL predictor generates pseudo-labels using the unlabeled data. The AL sampler employs a joint sampling strategy to select the most informative scenes. The oracle furnishes accurate labels for the sampled scenes, updating the labeled dataset which is then to retrain the AL predictor. The iterative nature of this process continues until the desired number of sampled scenes is attained.}   
    \label{fig:intro} 
\end{figure}

\subsubsection{Category Imbalance}
In AD datasets, there exists a discernible imbalance among various object classes (e.g., the quantity of cars greatly outweighs that of cyclists, as illustrated in Fig. \ref{fig:intro}). This imbalance poses a potential bias, as the detection learner may not sufficiently learn information from minority classes. How to mitigate the category bias via a balanced sampling scheme is important to detection accuracy.

\subsubsection{Scene Redundancy}
The limited availability of collection areas and sequential data gathering results in a notable similarity between scenes, including comparable positions and category distributions of traffic participants. How to select the scenario with minimal similarity is critical to pick sparse data while simultaneously guaranteeing high performance for the active learner.

\subsubsection{Scene Undercomplexity}
Real-world driving presents a multitude of intricate and ever-changing scenarios. An effective training dataset should encompass many diverse and challenging traffic scenes, such as mutual occlusion, low-resolution measurements, and poor illumination, etc. How to actively explore these valuable and diverse data is a key to maximizing the upper bound of 3D detection performance.

Previous research activities aimed at reducing category bias have primarily focused on repetitively copying or sampling the minority classes\cite{zhu2019class,chen2023revisiting}. However, they cannot be used in AL methods due to the lack of introducing unlabeled data. Typically, the traffic scene similarity is assessed through features extracted from models\cite{liang2022exploring,hekimoglu2022efficient}. Yet, these feature-based methods are contingent on specific model structures within the 3D detection framework, potentially compromising computational efficiency in AL frameworks due to the high-dimensional features. While several metrics, like statistical complexity\cite{liu2019towards,sadat2021diverse} and perception uncertainty\cite{choi2018uncertainty,feng2019deep,choi2021active}, have been proposed to gauge scene complexity. The former requires prior knowledge for setting weights across various statistical metrics, making it unsuitable for individual scene data selection. In contrast, the latter can quantify scene complexity arising from both scene data and deep neural network models.

In this paper, we propose a TSceneJAL framework designed to construct an optimal AD dataset by using a multi-stage sampling strategy to iteratively update the labeled dataset from unlabeled data. The main contributions of this work include:

\begin{itemize}
    \item {Developing a first-stage AL sampling strategy based on a category entropy metric to conditionally filter candidate traffic scenes, effectively alleviating object class bias;}
    \item {Developing a second-stage AL sampling strategy based on a scene similarity metric by modeling the traffic scenes with object graphs and measuring their graph distances, thereby diminishing scene redundancy;}
    \item {Developing a third-stage AL sampling strategy based on an uncertainty metric of mixture density network (MDN) to estimate the intrinsic stochasticity in both model and data, quantifying the complexity and diversity of the scenes.}
    \item Releasing the source code of our method at \url{https://github.com/ansonlcy/TSceneJAL}.
\end{itemize}

The remainder of this paper is organized as follows: Sec. \ref{sec:related} reviews related works in AL. Sec. \ref{sec:state_setup} states the framework setup and problem statement. Sec. \ref{sec:method} provides a detailed description of the AL sampler, including three evaluation metrics and the multi-stage sampling strategy. Sec. \ref{sec:experiment} presents experiments and analyses. Sec. \ref{sec:conclution} concludes this paper.
\section{Related Work}
\label{sec:related}
Active learning seeks to minimize the amount of labeled data required to achieve high information utility by strategically selecting the data instances to be labeled by the oracle \cite{krishnakumar2007active,feng2019deep,luo2023exploring}.
The data sampling strategy is the essence of active learning and can be mainly categorized into three types: distribution-based, complexity-based, and hybrid.
\subsection{Distribution-based Sampling}
The distribution based sampling strategies aim to select a subset of the dataset that faithfully mirrors the overall distribution. 
Various methods are employed to characterize sample distribution, including diversity and representativeness. Diversity-based methods\cite{liang2022exploring,sinha2019variational, hekimoglu2022efficient} prioritize samples with substantial differences in spatio-temporal distributions, traffic participant categories, and other factors, while they often require precise global coordinates and could hardly apply to non-sequential scenes, such as KITTI. 
Conversely, representative methods focus on encapsulating the intrinsic dataset structures. For instance, employing the K-means method to cluster sample points and then selecting high probability samples\cite{bodo2011active}, \cite{liao2022traffic} is a common approach. Nonetheless, clustering algorithms are sensitive to initial point selection, leading to low robustness. An alternative method, Coreset\cite{sener2017active}, assumes that AL is to minimize the coreset error and provides an approximate upper bound for the loss based on the feature distances. However, the high-dimensional features extracted from 3D detection models, usually coupled with irrelevant information like environmental details, would degrade AL efficiency. To address these issues, we propose an approach: initially selecting scenes using category entropy, followed by modeling scenes with directed graphs, which explicitly represent the distributions of traffic participant categories and positions, facilitating effective measurement of scene distances.

\subsection{Complexity-based Sampling}
Currently, the complexity evaluation metrics of AD datasets include statistical complexity\cite{liu2019towards,sadat2021diverse} and perception uncertainty\cite{gal2016dropout,wang2014new,joshi2009multi}. Statistical methods employ a set of metrics such as object quantity and density across categories, quantified through entropy, while those approaches are limited due to the need for certain prior knowledge (such as determining the weights of different metrics). Therefore, in AL methods, perception uncertainty is more prevalent as an evaluation metric for scene complexity. Perception uncertainty operates on the premise that samples with high uncertainty offer more information for models\cite{liu2022survey}. Typical uncertainty measurement methods include entropy uncertainty\cite{shannon2001mathematical}, \cite{wu2003probability}, confidence uncertainty\cite{wang2014new}, and BvSB uncertainty\cite{joshi2009multi}, \cite{netzer2011reading}. But these methods are primarily utilized in classification tasks, rather than assessing regression uncertainty.

Monte-Carlo Dropout\cite{gal2016dropout,feng2019deep, moses2022localization}, is an effective method for measuring regression uncertainty. However, its reliance on multiple forward propagations can impede efficiency. Ensemble methods\cite{lakshminarayanan2017simple,beluch2018power} offer an alternative but demand substantial resources since multiple models are trained concurrently. 
Additionally, Bayesian networks are explored by adding a Gaussian distribution over model parameters, optimizing weights, bias, and variance during training to estimate perception uncertainty\cite{choi2018uncertainty},\cite{choi2021active}. The drawback of this method is that a single Gaussian can only estimate aleatoric uncertainty (AU), ignoring epistemic uncertainty (EU). To address this, we develop a mixture density network (MDN) that capitalizes on the properties of the mixture Gaussians to accurately assess the scene complexity, enabling simultaneous estimation of both AU and EU in a single forward process.

\subsection{Hybrid Sampling}
In most cases, relying solely on individual selection strategies may not yield the most rational and informative samples. A straightforward approach\cite{hekimoglu2022efficient} is to directly combine multiple metrics through hyperparameters, but it often struggles to strike a balance between these metrics. The Badge\cite{ash2019deep} method leverages the maximum probability label of unlabeled data as pseudo-labels, computing gradient vectors, and utilizing K-means clustering to group samples. This method intelligently combines uncertainty and diversity without necessitating manual adjustment of hyperparameters. However, since regression tasks are challenging to provide pseudo-labels, this approach is not suitable for 3D object detection. A more generalized approach, Exploitation-Exploration\cite{yin2017deep,wu2022entropy}, operates in two stages. Initially, it selects samples with the maximum uncertainty and minimum redundancy, followed by the selection of samples with the maximum diversity. To utilize the assortment of metrics mentioned earlier, we extend this two-stage method and, similar to the Crb\cite{luo2023exploring} approach, develop a joint three-stage sampling strategy for active learning.

\section{Theoretical Foundation}
The AL goal is to build a labeled dataset $\mathcal{D}_S$ from raw data $\mathcal{D}_u$, while minimizing its empirical risk $\mathfrak{R}_T$ on a test set $\mathcal{D}_T$. The upper bound on empirical risk has been provided in \cite{luo2023exploring} as:
\begin{equation}
    \label{eq:total_risk}
    \begin{aligned}
        \mathfrak{R}_T[\ell(f;\omega)] \le \mathfrak{R}_S[\ell(f;\omega)]+\frac{1}{2}disc(\hat{\mathcal{D}}_S,\hat{\mathcal{D}}_T)+\lambda ^{*}+const ,
    \end{aligned}
\end{equation}
where $\mathfrak{R}_S$ is the empirical loss on the selected dataset $\mathcal{D}_S$. $\ell(f;\omega)$ is the loss function of model $f$ with parameters $\omega$. $\lambda^{*}$ is the empirical loss of the optimal model. $const$ remains a constant value under the condition of a fixed size of the selected dataset.  $\hat{\mathcal{D}}_S$ and $\hat{\mathcal{D}}_T$ denote the empirical distributions of $\mathcal{D}_S$ and $\mathcal{D}_T$, respectively. $disc(\cdot,\cdot)$ is the function used to measure the disparity between distributions.

Minimizing $\mathfrak{R}_T$ can be achieved by minimizing its upper bound. Therefore, according to Eq. (\ref{eq:total_risk}), the optimization objective of active learning is:
\begin{equation}
    \label{eq:ch1_get_ds}
    \begin{aligned}
        \mathcal{D}^{*}_{S}=\mathop{\arg\min}_{\mathcal{D}_{S}\subset \mathcal{D}_u} \{  disc(\hat{\mathcal{D}}_{S},\hat{\mathcal{D}}_{T}) + \mathfrak{R}_S[\ell(f;\omega)]  \},
    \end{aligned}
\end{equation}
{where $\mathcal{D}^{*}_{S}$ represents the optimal dataset constructed from raw data $\mathcal{D}_u$.}
{
Usually a training algorithm needs to minimize the empirical risk $\mathfrak{R}_S$ in Eq. (\ref{eq:ch1_get_ds}) as well, however our active learning scheme focuses on minimizing the first term according to the zero-training error assumption \cite{sener2017active}.
In this paper, we defined the term $disc(\cdot,\cdot)$ in Eq. (\ref{eq:ch1_get_ds}) as the following three parts: 
1) data balance, reflecting the disparity in the distribution of selected categories between $\mathcal{D}_S$ and $\mathcal{D}_T$;
2) data diversity, indicating the discrepancy in the scene similarity distribution between $\mathcal{D}_S$ and $\mathcal{D}_T$;
3) data complexity, referring to the difference in the intricacy distribution between $\mathcal{D}_S$ and $\mathcal{D}_T$,
\begin{equation}
    \label{eq:ch1_final_dis}
    \begin{aligned}
        \mathcal{D}^{*}_{S} \approx \mathop{\arg\min}_{\mathcal{D}_{S}\subset \mathcal{D}_u}  \{ \underbrace{D_{KL}(P_{Y_S},P_{Y_T})}_{balance} +\\ \underbrace{D_{KL}(P_{G_S},P_{G_T})}_{diversity} + \underbrace{D_{KL}(P_{\mathcal{D}_S},P_{\mathcal{D}_T})}_{complexity} \},
    \end{aligned}
\end{equation}
{where $D_{KL}(\cdot, \cdot)$ stands for the Kullback-Leibler divergence between two distributions; $P_{Y_S}$ and $P_{Y_T}$ denote the label category distributions of $\mathcal{D}_S$ and $\mathcal{D}_T$, respectively, with $P_{Y_T}$ assumed to follow a uniform distribution; $P_{G_S}$ and $P_{G_T}$ are the distributions of data similarity, with $P_{G_T}$ following a Gaussian distribution; the target data distribution $P_{D_T}$  is assumed to be uniform.}
}

Based on the Eq. (\ref{eq:ch1_final_dis}), three data metrics are proposed in this paper, corresponding to balance, {diversity}, and complexity. The optimal AD dataset will be constructed through a multi-stage strategy, progressively utilizing three data metrics:

\subsubsection{Balance} To address the issue of imbalance within the dataset, we consider the first term of Eq. (\ref{eq:ch1_final_dis}):
\begin{equation}
    \label{eq:ds_1}
    \begin{aligned}
          & \mathcal{D}_{S_1}^{*}=\mathop{\arg\min}_{\mathcal{D}_{S_1}\subset \mathcal{D}_{u}}D_{KL}(P_{Y_{S_1}}\left |  \right |P_{Y_{T}})= \\ & \mathop{\arg\min}_{\mathcal{D}_{S_1}\subset \mathcal{D}_{u}}{\left [ \sum_{c=1}^{C}{P_{Y_{S_1}}(c)\log{P_{Y_{S_1}}(c)}}-\sum_{c=1}^C{P_{Y_{S_1}}(c)\log{P_{Y_T}}(c)}\right ]},
    \end{aligned}
\end{equation}
where {$\mathcal{D}_{S_1}$ is the dataset constructed from $\mathcal{D}_{u}$ in the first stage.} $c\in \{ 1,2,..,C \}$ is the category. Given the assumption that $P_{Y_T}$ follows a uniform distribution, we have $\log{P_{Y_T}}(c)=-\log{C}$, and $\sum_{c=1}^C{P_{Y_{S_1}}(c)}=1$. Consequently, Eq. (\ref{eq:ds_1}) can be expressed as follows:
\begin{equation}
    \label{eq:ds_1_2}
    \begin{aligned}
          & \mathcal{D}_{S_1}^{*} = \mathop{\arg\max}_{\mathcal{D}_{S_1}\subset \mathcal{D}_{u}} H_{d}(Y_{S_1}),
    \end{aligned}
\end{equation}
{where $H_{d}(\cdot)$ is the function for calculating the entropy of discrete variables, $H_{d}(Y_{S_1})= -\sum_{c=1}^{C}{P_{Y_{S_1}}(c)\log{P_{Y_{S_1}}(c)}}$.}

According to Eq. (\ref{eq:ds_1_2}), the construction of the optimal dataset $\mathcal{D}_{S_1}^{*}$ entails maximizing the entropy across various category quantities. Then, a greedy strategy can be employed to progressively select scenes from the dataset, prioritizing those with the highest category entropy:
\begin{equation}
    \label{eq:ds_1_3}
    \begin{aligned}
          & \mathcal{P}^{*} = \mathop{\arg\max}_{\mathcal{P}\subset \mathcal{D}_{u}} Entropy(\mathcal{P}),
    \end{aligned}
\end{equation}
{where $\mathcal{P}$ is a scene from the raw dataset $\mathcal{D}_{u}$.}

Based on the derivation above, we proposes a \textbf{category entropy metric} for scene sampling in Sec. \ref{sec:cate}.

\subsubsection{Diversity} To address the issue of redundancy within the dataset, we consider the second term of Eq. (\ref{eq:ch1_final_dis}):
\begin{equation}
    \label{eq:ds_2}
    \begin{aligned}
         & \mathcal{D}_{S_2}^{*}=\mathop{\arg\min}_{\mathcal{D}_{S_2}\subset \mathcal{D}_{S_1}}D_{KL}(P_{G_{S_2}}\left |  \right |P_{G_{T}})=\\ & \mathop{\arg\min}_{\mathcal{D}_{S_2}\subset \mathcal{D}_{S_1}}\log{\frac{\sigma_T}{\sigma_{S_2}}}+\frac{\sigma^2_{S_2}+(\mu_{S_2}-\mu_{T})}{2\sigma^2_{T}}-\frac{1}{2} \approx FS(G_{S_1}),
    \end{aligned}
\end{equation}
where {$\mathcal{D}_{S_2}$ is the dataset constructed from $\mathcal{D}_{S_1}$ in the second stage.} $G_{S_2}=\{Similarity(\mathcal{P}^i,\mathcal{P}^j)\}_{ \mathcal{P}^i,\mathcal{P}^j \in \mathcal{D}_{S_2} }$ is the set of similarity within $\mathcal{D}_{S_2}$. Both distribution $P_{G_{S_2}}$ and $P_{G_{T}}$ are assumed to conform to Gaussian distributions, with $\mu$ and $\sigma$ being their mean and variance, respectively. $FS(\cdot)$ denotes the farthest sampling algorithm, designed to prioritize the selection of more diverse samples. This algorithm necessitates obtaining the similarity between scenes. To tackle this requirement, we proposes the \textbf{{scene} similarity metric} in Sec. \ref{sec:similarity}.

{
\subsubsection{Complexity} To address the issue of under-complexity within the selected dataset, we consider the third term of Eq. (\ref{eq:ch1_final_dis}). 
Similar to Eq. (\ref{eq:ds_1_2}), we use the entropy for evaluating KL-Divergence with the uniform distribution $P_{\mathcal{D}_{T}}$ as a metric for complexity:
\begin{equation}
    \label{eq:ds_3}
    \begin{aligned}
        \mathcal{D}_{S_3}^{*} &= \mathop{\arg\min}_{\mathcal{D}_{S_3}\subset \mathcal{D}_{S_2}} D_{KL}(P_{\mathcal{D}_{S_3}} || P_{\mathcal{D}_T}) \\
        &= \mathop{\arg\max}_{\mathcal{D}_{S_3}\subset \mathcal{D}_{S_2}} H(\mathcal{D}_{S_3}) \\
        &= \mathop{\arg\max}_{\mathcal{D}_{S_3}\subset \mathcal{D}_{S_2}} [H(\mathcal{D}_{S_3} | \omega) + H(\omega) - H(\omega | \mathcal{D}_{S_3})] ,
    \end{aligned}
\end{equation}
where $\mathcal{D}_{S_3}$ is the dataset constructed from $\mathcal{D}_{S_2}$ in this stage.
While entropy $H(w)$ represents the complexity of the prior on the model parameters, which is typically treated as a constant once the distribution of the parameters is established.
Besides, $H(\mathcal{D}_{S_3}|\omega)$ stands for the complexity of $\mathcal{D}_{S_3}$ with parameter $\omega$:
\begin{equation}
H(\mathcal{D}_{S_3}|\omega) = \mathbb{E}_{\mathcal{D}_{S_3}\sim p(\mathcal{D}_{S_3}|\omega)}[-\log p(\mathcal{D}_{S_3}|\omega)].
\end{equation}
Since the model trained on the selected data in previous AL iterations tends to yield stable outputs,
$H(\omega|D_{S_3}) \ll H(\mathcal{D}_{S_3}| \omega)$. We focus on the last term in Eq. (\ref{eq:ds_3}):
\begin{equation}
    \begin{aligned}
        \mathcal{D}_{S_3}^{*} 
        &\approx \mathop{\arg\min}_{\mathcal{D}_{S_3}\subset \mathcal{D}_{S_2}} H(\omega|\mathcal{D}_{S_3}).
    \end{aligned}
\end{equation}
Thus, constructing the optimal dataset $\mathcal{D}_{S_3}^{*}$ involves minimizing its conditional entropy. Let $\mathcal{P}^*$ denote the data in $\mathcal{D}_{S_3}^{*}$ and the method for {seeking data point} $\mathcal{P}^*$ is by \cite{houlsby2011bayesian}:
\begin{equation}
    \label{eq:ds_3_2}
    \begin{aligned}
         \mathcal{P}^{*} &=\mathop{\arg\max}_{\mathcal{P}\in \mathcal{D}_{S_2}}\left[ H(\omega | \mathcal{D}_{l})  - E_{y\sim p(y|\mathcal{P},\mathcal{D}_{l})}[ H(\omega|\mathcal{D}_{l},\mathcal{P},y) ]\right] \\
         &= \mathop{\arg\max}_{\mathcal{P}\in \mathcal{D}_{S_2}}\left[H(y|\mathcal{P},\mathcal{D}_{l})  - E_{\omega \sim p(\omega |\mathcal{D}_{l})} H(y|\mathcal{P},\omega) \right] \\
         &\approx \mathop{\arg\max}_{\mathcal{P}\in \mathcal{D}_{S_2}} Uncertainty(\mathcal{P}),
    \end{aligned}
\end{equation}
where $y$ is the prediction of data input $\mathcal{P}$, and {$\mathcal{D}_{l}$ is the labeled data for model training}.
}
As indicated by Eq. (\ref{eq:ds_3_2}), constructing the optimal subset entails selecting scenes with the highest uncertainty from the pool of candidate scenes. In this paper, we propose a \textbf{perception uncertainty metric} for scenes in Sec. \ref{sec:uncertainty}.

\section{System setup and Problem Statement}
\label{sec:state_setup}
\subsection{System Setup}
\label{sec:setup}
As shown in Fig. \ref{fig:system}, our proposed TSceneJAL framework comprises two main components: a basic AL predictor and an AL sampler based on the category entropy, scene similarity and perception uncertainty metrics. 
 \subsubsection{\textbf{AL predictor}}
 The left part of the Fig. \ref{fig:system} illustrates the AL predictor $f_m(\omega_r;\cdot)$, where $\omega_r$ is the model weight after training with the $r$-th round update of the labeled data. Unlike traditional processes, we transform the predictor's regression head into a mixture density network (MDN), as detailed in Sec. \ref{sec:uncertainty}. Consequently, for a given input scene $\mathcal{P}^t$, the AL predictor simultaneously yields classification $\hat{\mathcal{C}}^t$ and MDN result $\hat{\mathcal{M}}^t$. Furthermore, box regression results $\hat{\mathcal{B}}^t$ can be derived from $\hat{\mathcal{M}}^t$. 
 \subsubsection{\textbf{AL sampler}}
\begin{figure*}  
    \centering     
    \includegraphics[width=1.0\linewidth]{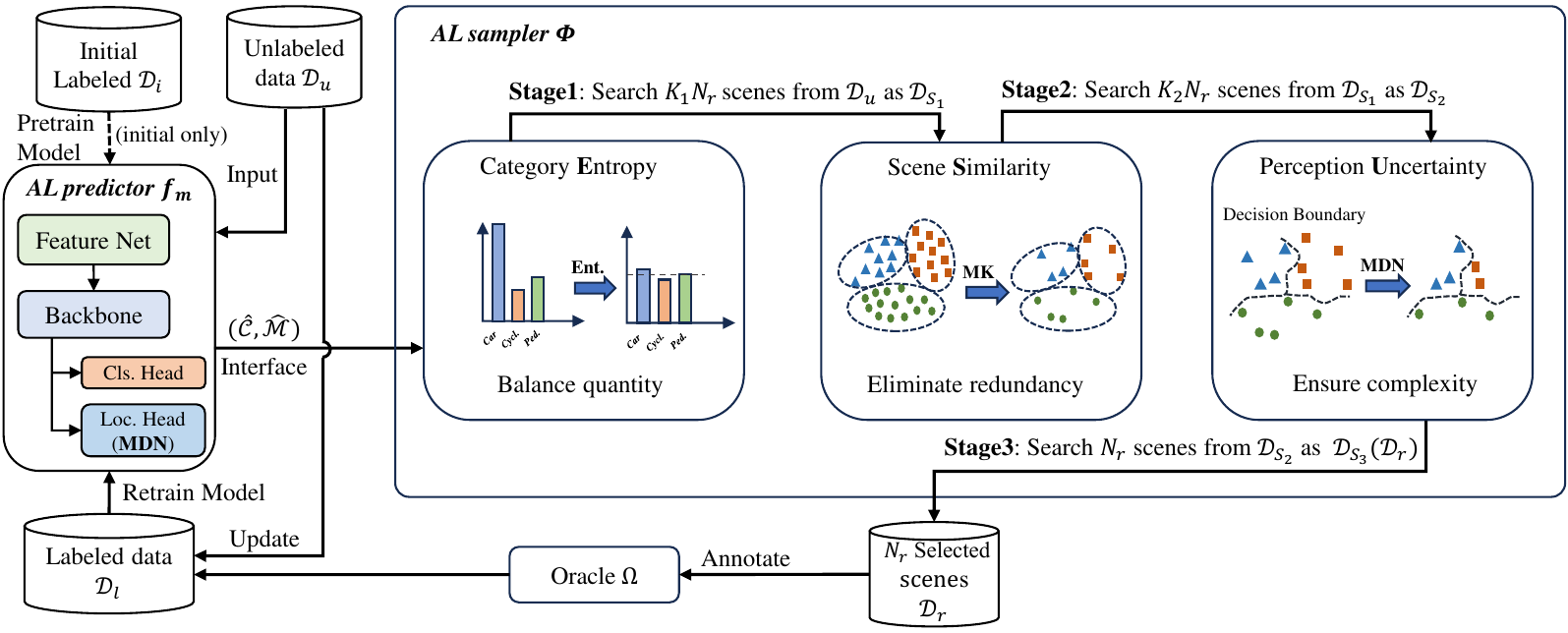}     
    \caption{An illustration of our TSceneJAL framework. The AL predictor $f_m$ is transformed by integrating a MDN and pre-trained using the initial labeled dataset $\mathcal{D}_i$. The trained AL predictor $f_m$ generates pseudo-labels and MDN predictions, $(\hat{\mathcal{C}},\hat{\mathcal{M}})$, from the unlabeled dataset $\mathcal{D}_u$. The AL sampler incorporates three metrics, namely category entropy (\ref{sec:cate}), scene similarity (\ref{sec:similarity}) and perception uncertainty (\ref{sec:uncertainty}), to evaluate the data. A three-stage hybrid sampling strategy (\ref{sec:selection}) is then used for data selection, resulting in a set of $N_r$ required scenes. These selected scenes are annotated by the oracle $\Omega$ and subsequently used for model retraining.}   
    \label{fig:system} 
\end{figure*}

The sampler $\phi$ is the core of the AL framework. In the $r$-th iteration of AL, the sampler selects $N_r$ scenes, denoted as $\mathcal{D}_r$, from the unlabeled dataset $\mathcal{D}_u$. These selected scenes are then labeled by the annotator $\Omega$ and appended to the updated labeled pool, $\mathcal{D}_l$. The AL process continues until the annotation budget $B$ is depleted, \textit{i.e.}, $\sum_1^{R}N_r\geq{B}$, where $R$ is the total number of AL iterations. Specifically, the AL sampler incorporates three evaluation metrics and a joint three-stage selection strategy.
\textbf{(a) Evaluation metrics}: Within our TSceneJAL framework, three metrics are employed for scene evaluation. One is the category entropy metric, $E(\cdot)$, utilized to compute the entropy of object occurrence probabilities in scenes, mitigating category imbalance in the dataset, as detailed in Sec. \ref{sec:cate}. Another one is the similarity metric, $S(\cdot, \cdot)$, employed to measure the distance between two scenes, where the objects in each scene are linked as a graph representation and then their similarity is quantified by the marginalized kernel algorithm \cite{kashima2003marginalized}. The last metric, the uncertainty metric $U(\cdot)$, estimates uncertainy in both model and data via an MDN network, which is based on the assumption that selecting  scenes with high uncertainty can enhance the complexity of the data, as detailed in Sec. \ref{sec:uncertainty}.
 \textbf{(b) Multi-stage joint selection strategy}: Based on the evaluation metrics, a joint sampling strategy is devised for data selection. The selection process across different stages is presented in Fig. \ref{fig:system}. Similar to \cite{luo2023exploring}, we employ a three-stage sampling strategy, progressively utilizing each metric for data selection. Further details are provided in Sec. \ref{sec:selection}.
 
\subsection{Problem Statement}
Hence, our focus in this work revolves around addressing the following problems: 

\begin{itemize}
    \item How to define the category entropy metric $E^t(\hat{\mathcal{C}}^t)$ for scene $\mathcal{P}^t$, aiding in the selection of those scenes to mitigate the impact of object class bias within the dataset;
    \item How to define the scene similarity metric $S^{i,j}(\mathcal{P}^i, \mathcal{P}^j)$ between scenes $\mathcal{P}^i$ and $\mathcal{P}^j$, which can help to reduce dataset redundancy;
    \item How to define the perceptual uncertainty metric $U^t(\hat{\mathcal{M}}^t)$ for scene $\mathcal{P}^t$, intending to maintain the complexity of scenes within the dataset. 
    \item How to effectively utilize the above three metrics to select scenes $\mathcal{D}_r$ from $\mathcal{D}_u$, achieving a balanced relationship among them.
\end{itemize}

\section{Evaluation Metrics}
\label{sec:method}

\subsection{Category Entropy Metric}
\label{sec:cate}
Due to the inherent characteristics of real traffic scenes, AD datasets often manifest significant imbalances in the numbers of different traffic participants during collection (such as in KITTI dataset, the number of car instances far exceeds that of pedestrians and cyclists). 
This category bias results in the predictor lacking the requisite capability to effectively learn features of minority classes. 

For the traffic scene of $t$-th frame, $\mathcal{P}^t$, its object classification predicted by $f_m$ is $\hat{\mathcal{C}}^t=\{ \hat{c}^{t}_{i}\}_{i\in[\hat{N}^t]}$, where $\hat{c}^{t}_{i}$ is the class label for the $i$-th bounding box, and $\hat{N}^t$ is the predicted number of boxes in $\mathcal{P}^t$. Then the category entropy of the current scene, $E^t(\cdot)$, can be calculated by:
\begin{equation}
    \label{eq:cate_entropy}
    \begin{aligned}
       E^t(\hat{\mathcal{C}}^t)=-\sum^C_{c=1}{\hat{p}^{t}_{c}\log{(\hat{p}^{t}_{c}+\zeta )}},
    \end{aligned}
\end{equation}
where $\hat{p}^{t}_{c}$ is the proportion of objects predicted as class $c$ and $C$ is all classes specified in the traffic dataset, $\zeta$ is a small constant value to ensure the calculation stability. And $\hat{p}^{t}_{c}$ can be obtained by:
\begin{equation}
    \label{eq:cate_prob}
    \begin{aligned}
       \hat{p}^{t}_{c}=\frac{\sum_{i=1}^{\hat{N}_t}{\delta(\hat{c}^{t}_{i},c,\tau)}}{\sum_{c=1}^C\sum_{i=1}^{\hat{N}_t}{\delta(\hat{c}^{t}_{i},c,\tau)}},
    \end{aligned}
\end{equation}
\begin{equation}
    \label{eq:delta_func}
    \begin{aligned}
        \delta(\hat{c}^{t}_{i},c,\tau)=
            \left\{\begin{matrix} 1
            & \hat{c}^{t}_{i}=c \wedge (conf.(\hat{c}^{t}_{i})\ge \tau)  \\
            0 & \hat{c}^{t}_{i} \ne c \vee (conf.(\hat{c}^{t}_{i})<\tau) 
        \end{matrix}\right. ,
    \end{aligned}
\end{equation}
{where $conf.(\cdot)$ is the confidence of classification predicted result by the model $f_m$. Applying a threshold helps filter out ineffective predictions, thereby enhancing computation accuracy. The value of threshold $\tau$ is set to 0.3 in this paper, and we conduct experiments on its value in the following sections.}


\subsection{Scene Similarity Metric}
\label{sec:similarity}
Most AD datasets are gathered in a continuous manner, resulting in neighboring scenes exhibiting striking similarities. Consequently, such datasets often carry a significant redundancy. When the amount of data is fixed, reducing redundancy can increase data diversity, which in turn enhances the model's generalization ability and improves its performance on the test set. To mitigate redundancy and bolster the diversity of selected data, we adopt a similarity measurement grounded in scene-directed graphs. In this section, we will outline the process of representing a traffic scene as a graph and using the marginalized kernel algorithm to compute scene similarity. 

\subsubsection{\textbf{Graph representation of a traffic scene}}
\begin{figure*}  
    \centering     
    \includegraphics[width=1.0\linewidth]{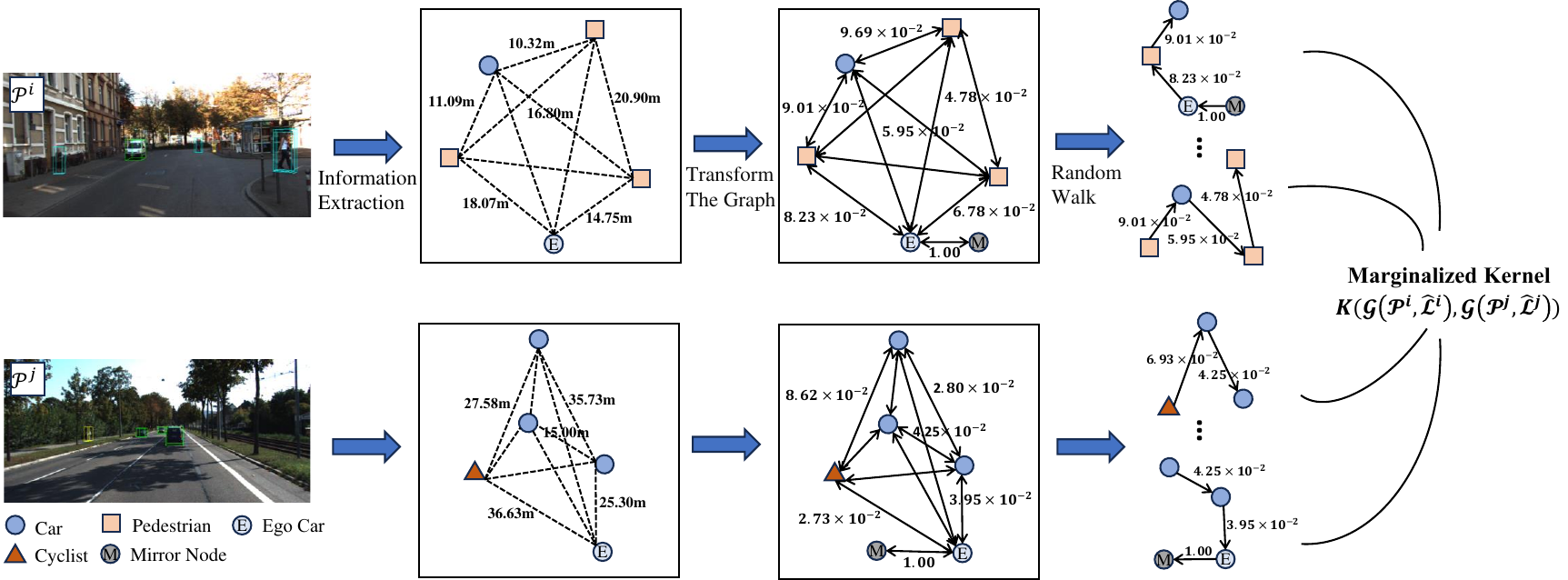}     
    \caption{An illustration of graph representation of scenes and scene similarity estimation. The process comprises several steps: 1) Information extraction from point clouds using the AL predictor to obtain predicted target boxes and labels. 2) Construction of a fully connected undirected graph structure with box categories as nodes and distances as edges, including the addition of a node for the ego vehicle. 3) Transformation of the undirected graph into a directed graph, updating edge weights, and introducing mirrored nodes. 4) Utilization of random walks to generate multiple subgraphs. 5) Calculation of similarity between graphs using Eq. (\ref{eq:simi_frame}).}   
    \label{fig:frame_graph} 
\end{figure*}

{A traffic scene consists of the static environment (\textit{e.g.}, tree and building) and the dynamic objects (\textit{e.g.}, vehicle and pedestrian), where the variation of the dynamic ones contributes to the difference between scenes.}
For a traffic scene $\mathcal{P}^t$, its classification $\hat{\mathcal{C}}^t=\{\hat{c}^{t}_{i}\}_{i\in[\hat{N}^t]}$ and regression $\hat{\mathcal{B}}^t= \{\hat{b}^{t}_{i}\}_{i\in[\hat{N}^t]}$ can be obtained from the AL predictor. A fully connected undirected graph can be constructed as $\mathcal{G}(\mathcal{P}^t)\rightarrow G^t=\{\{v^t_i\},\{e^t_{ij}\} \}_{i,j\in[\hat{N}^t+1]}$, where $\mathcal{G}(\cdot)$ is the graph generation process. Each node, $v^t_i=\hat{c}^{t}_{i}$, represents a dynamic object, including the ego vehicle. 
The edge is defined as $e^t_{ij}=1 / dis(\hat{b}^{t}_{i}, \hat{b}^{t}_{j})$, where $dis(\cdot, \cdot)$ denotes the Euclidean distance. The calculation of distance relies on 3D coordinate information, so the metrics in this section are more suitable for 3D object detection tasks. To enable the marginalized kernel algorithm \cite{kashima2003marginalized} to calculate similarity between directed graphs, $G^t$ is transformed into directed graphs by replacing the undirected edges $e^t_{ij}$ with bidirectional edges $\overrightarrow{e^t_{ij}}$ and $\overrightarrow{e^t_{ji}}$. Meanwhile, in cases where there are no movable objects except the ego vehicle, a mirror node is introduced. This node is solely connected to the ego vehicle with an edge weight of 1. Fig. \ref{fig:frame_graph} shows the conversion of a scene with various objects into a directed graph.

\subsubsection{\textbf{Similarity based on marginalize kernel algorithm}}
The marginalized kernel algorithm \cite{kashima2003marginalized} is a type of graph kernel employed for evaluating the similarity between various directed graphs. 
Given a graph, it generates a series of paths, denoted as $\boldsymbol{h}=(h_1,h_2,...,h_\ell)$, though random walks, where each path $h_\ell$ encompasses a succession of $\ell$ steps within the graph space. 
The marginalized kernel defines the overall similarity between graphs $G$ and graphs $G^{'}$ as the expected value of the similarity between all pairs of paths on both graphs simultaneously. The similarity is calculated by:
\begin{equation}
    \label{eq:mk}
    \begin{aligned}
        K( {G,G^{'}} ) = {\sum_{\ell}^{\infty}\sum_{\boldsymbol{h}}{{\sum_{\boldsymbol{h}^{'}}{p( \boldsymbol{h} | G )p( \boldsymbol{h}^{'} | G^{'} )K_{\textit{z}}( \textit{z},\textit{z}^{'} )}}}},
    \end{aligned}
\end{equation}
where {$p( \boldsymbol{h} | G )$ represents the probability of obtaining path $\boldsymbol{h}$ from graph $G$ through a random walk.} $\textit{z}=(G,\boldsymbol{h})$, $K_{\textit{z}}( \textit{z},\textit{z}^{'} )$ is the joint kernel function, which can be obtained as follows:
\begin{equation}
    \label{eq:kernel}
    \begin{aligned}
    K_{\textit{z}}\left( {\textit{z},\textit{z}^{'} }\right) = \left\{ \begin{matrix}
{0, \quad ( \ell \neq \ell^{'} )} \\
{K( {v_{h_{1}},v_{h_{1}^{'}}^{'}} ){\prod_{i = 2}^{{\ell}}{K( {e_{h_{i - 1}h_{i}},e_{h_{i - 1}^{'}h_{i}^{'}}^{'}} )}}}\\ 
{\times K( {v_{h_{i}},v_{h_{i}^{'}}^{'}} ), \quad (\ell = \ell^{'})} \\
\end{matrix} \right.
    \end{aligned}
\end{equation}
where the kernel function between nodes is defined as $K(v,v^{'})=\xi(v=v^{'})/2$, while the kernel function between edges is given as $K(e,e^{'})=\exp{({- \| e-e^{'}  \| }/{2\sigma^2})}$. The value of $\xi(\cdot)$ is 1 if its argument holds true, and 0 otherwise. 
The similarity between two scenes $\mathcal{P}^i$ and $\mathcal{P}^j$ can be expressed as:
\begin{equation}
    \label{eq:simi_frame}
    \begin{aligned}
        S^{i,j}(\mathcal{P}^i, \mathcal{P}^j)=K(G^i,G^j).
    \end{aligned}
\end{equation}

To mitigate data redundancy, we adopt the farthest point sampling scheme, extracting scenes with the greatest dissimilarity to the current labeled data during each iteration. Additionally, we conduct sampling of initial points to ensure the stability. The details are shown in Algorithm \ref{alg:farthest_sampling_alg}.

\begin{algorithm}[h]
\caption{The Farthest Sampling Algorithm}
\label{alg:farthest_sampling_alg}
\begin{algorithmic}[1]
\REQUIRE ~~\\ 
$\mathcal{D}_{S_1}$: Unlabeled data in the first stage \\
$K_1N_r$: The number of samples in $\mathcal{D}_{S_1}$\\
$K_2N_r$: The number of scenes to be selected\\
\ENSURE ~~\\
$\mathcal{D}_{S_2}$: The selected scenes
\STATE $\mathcal{D}_{S_2} \leftarrow \emptyset$
\FOR{each $\mathcal{P}^t \in \mathcal{D}_{S_1}$}
    \STATE $sim^t(\mathcal{P}^t)=\sum\limits_{k=1}^{K_1N_r}S^{t,k}(\mathcal{P}^t,\mathcal{P}^k)$
    \COMMENT{similarity by Eq. (\ref{eq:simi_frame})}
\ENDFOR
\STATE $\mathcal{P}^g=\mathop{\arg\max}\limits_{\mathcal{P}^t \in \mathcal{D}_{S_1}}sim^t(\mathcal{P}^t) $
\STATE $\mathcal{P}^0=\mathop{\arg\min}\limits_{\mathcal{P}^t \in \mathcal{D}_{S_1}/\mathcal{P}^g}S^{t,g}(\mathcal{P}^t,\mathcal{P}^g)$ \COMMENT{determine initial scene}
\STATE $\mathcal{D}_{S_2} \leftarrow \mathcal{D}_{S_2} \cup \mathcal{P}^0$
\STATE $\mathcal{D}_{S_1} \leftarrow \mathcal{D}_{S_1} / \mathcal{P}^0$
\WHILE{$\left | \mathcal{D}_{S_2} \right | < K_2N_r$}
    \STATE ${\mathcal{P}^s}^*=\mathop{\arg\max}\limits_{\mathcal{P}^s \in \mathcal{D}_{S_1}} ( \mathop{\min}\limits_{\mathcal{P}^j \in \mathcal{D}_{S_2}}(1-S^{s,j}(\mathcal{P}^s,\mathcal{P}^j)) ) $ \COMMENT{greedy sampling}
    \STATE $\mathcal{D}_{S_2} \leftarrow \mathcal{D}_{S_2} \cup {\mathcal{P}^s}^*$
    \STATE $\mathcal{D}_{S_1} \leftarrow \mathcal{D}_{S_1} / {\mathcal{P}^s}^*$
\ENDWHILE
\STATE \textbf{return} $\mathcal{D}_{S_2}$

\end{algorithmic}
\end{algorithm}

\subsection{Perceptual Uncertainty Metric}
\label{sec:uncertainty}
{
Modern deep learning methods for uncertainty estimation\cite{kendall2017uncertainties} typically focus on two key sources: epistemic uncertainty (EU) and aleatoric uncertainty (AU). EU arises from uncertainty in the model parameters, often captured by a prior distribution, offering insight into the model's uncertainty. In contrast, AU reflects inherent randomness in the data, represented by a distribution over the model's outputs.
}
In this section, we delve into the construction of the  mixture density network (MDN), integrating both types of perception uncertainty to enhance evaluation of scene complexity.

\begin{figure}  
    \centering     
    \includegraphics[width=1.0\linewidth]{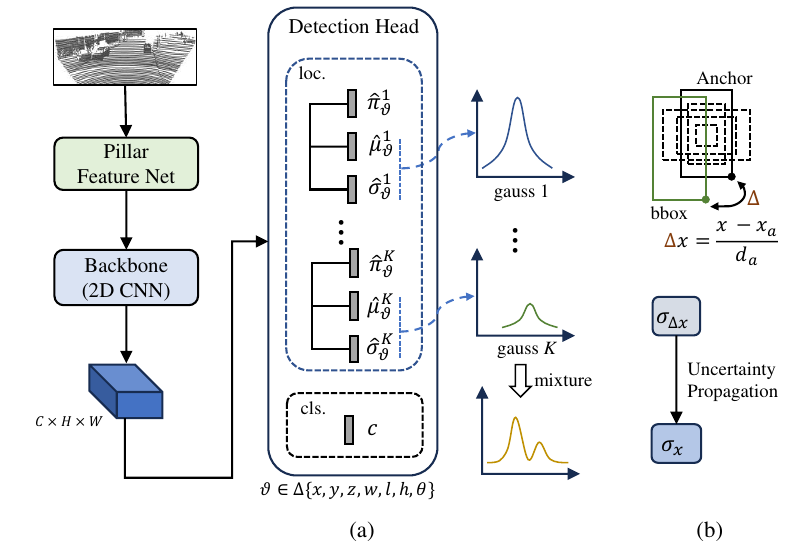}     
    \caption{An illustration of the MDN and correponding uncertainy propagation. (a) The MDN's structure, enhancing the regression head of the PointPillar model by modeling the output as a mixture of multiple Gaussian distributions, while the classification head remains unaltered. (b) Given its anchor-based method, the model predicts the residual with respect to the anchor; thus, the uncertainty obtained by the MDN reflects the uncertainty associated with the residual.}   
    \label{fig:mdn} 
\end{figure}

The PointPillars\cite{lang2019pointpillars} served as the baseline for constructing the MDN. The MDN output is denoted as $\hat{\mathcal{M}}^t=\{ \hat{\pi}^k_{\vartheta},\hat{\mu}^k_{\vartheta},\hat{\sigma}^k_{\vartheta}\}$, representing the weights ($\hat{\pi}^k_{\vartheta}$), means ($\hat{\mu}^k_\vartheta $), and variances ($\hat{\sigma}^k_\vartheta $), respectively. Here, $\vartheta \in \{ \triangle x,\triangle y,\triangle z,\triangle w,\triangle h,\triangle l,\triangle \theta \} $ stands for the 
residual of the predicted 3D bounding box $\hat{b}_i^{t}$, and $k\in\{1,\dots,K\}$ represents the $k$-th Gaussian distribution. The output subjects to the following constraints: $\hat{\pi}_{\vartheta}^{k} = e^{{\hat{\pi}}_{\vartheta}^{k}}/\sum_{j = 1}^{K}e^{{\hat{\pi}}_{\vartheta}^{j}}$ and ${\hat{\sigma}}_{\vartheta}^{k} = sigmoid\left( {\hat{\sigma}}_{\vartheta}^{k} \right)$. As shown in Fig. \ref{fig:mdn}(a), the network produces a mixture of $K$ Gaussian distributions for each residual element, enabling the calculation of uncertainty. 
The loss function for the MDN is derive from the negative log-likelihood function as:
\begin{equation}
    \label{eq:mdn_loss}
    \begin{aligned}
       & L_{loc}\left(\hat{b}^t_{i},b^t_{i} \right) =\\& - \frac{1}{\hat{N}^t}{\sum_{i = 1}^{\hat{N}^t}{{\sum_{\vartheta}{~ln}}{\sum_{k = 1}^{K}{\hat{\pi}_{\vartheta}^{k}\left(\hat{b}^t_{i} \right)\mathcal{N}\left( b^t_{i} \middle| \hat{\mu}_{\vartheta}^{k}\left( \hat{b}^t_{i} \right),\hat{\sigma}_{\vartheta}^{k}\left( \hat{b}^t_{i} \right) \right)}}}},
    \end{aligned}
\end{equation}
where $\hat{b}^t_{i}$ is the $i$-th predicted object in the current point cloud $\mathcal{P}^t$, $b^t_{i}$ is its ground truth. $\hat{\mu}_{\vartheta}^{k}(\cdot)$ represents the mean of the MDN corresponding to a specific target, and $\hat{\sigma}_{\vartheta}^{k}(\cdot)$ as well as $\hat{\pi}_{\vartheta}^{k}(\cdot)$ later in the paper follow the same notation. $\mathcal{N}(\cdot)$ is the Gaussian distribution.
The sizes for regressing the bounding box can be given as: 
\begin{equation}
    \label{eq:mu_result}
    \begin{aligned}
       \hat{b}^t_{i\vartheta}\equiv \hat{\mu}^t_{i\vartheta} &= \sum_{k = 1}^{K}{\hat{\pi}_{\vartheta}^{k}(\hat{b}^t_{i})\hat{\mu}_{\vartheta}^{k}(\hat{b}^t_{i})}.
    \end{aligned}
\end{equation}
Referring to \cite{choi2018uncertainty}, the two uncertainties can be calculated by:
\begin{equation}
    \label{eq:var_1}
    \begin{aligned}
       \left [ \hat{\sigma}^t_{i\vartheta}\right ]_{au} = \sum_{k = 1}^{K}{\hat{\pi}_{\vartheta}^{k}(\hat{b}^t_{i})\hat{\sigma}_{\vartheta}^{k}(\hat{b}^t_{i})},
    \end{aligned}
\end{equation}
\begin{equation}
    \label{eq:var_2}
    \begin{aligned}
       \left [ \hat{\sigma}^t_{i\vartheta} \right ]_{eu} = \sum_{k = 1}^{K}{\hat{\pi}_{\vartheta}^{k}(\hat{b}^t_{i})\left\| {\hat{\mu}_{\vartheta}^{k}(\hat{b}^t_{i}) - {\hat{\mu}^t_{i\vartheta}}} \right\|^{2}},
    \end{aligned}
\end{equation}
where $\left [ \hat{\sigma}^t_{i\vartheta}\right ]_{au}$ and $\left [ \hat{\sigma}^t_{i\vartheta} \right ]_{eu}$ represent the AU and EU of the prediction residual $\vartheta$ for the $i$-th object in the scene $\mathcal{P}^t$, respectively.

Since PointPillar operates as an anchor-based model, as shown in Fig. \ref{fig:mdn} (b), it actually predicts the residuals relative to the anchor. Consequently, in Eq. (\ref{eq:var_1})-(\ref{eq:var_2}), the computed variance/uncertainty represents solely the variance of the residuals. To calculate the object uncertainty, a propagation calculation is required as follows:


\begin{equation}
    \label{eq:final_uncer}
    \begin{aligned}
        \hat{\sigma}_{ix}&=(d^a)^2\hat{\sigma}_{i\Delta x},\hat{\sigma}_{iy}=(d^a)^2\hat{\sigma}_{i\Delta y},\hat{\sigma}_{iz}=(h^a)^2\hat{\sigma}_{i\Delta z} \\
        \hat{\sigma}_{iw}& \approx \hat{\mu}_{i\Delta w}^2\hat{\sigma}_{i\Delta w},\hat{\sigma}_{il} \approx \hat{\mu}_{i\Delta l}^2\hat{\sigma}_{i\Delta l},\hat{\sigma}_{ih}\approx \hat{\mu}_{i\Delta h}^2\hat{\sigma}_{i\Delta h} \\
        \hat{\sigma}_{i\theta}& \approx sec^2(\hat{\mu}_{i\Delta \theta})\hat{\sigma}_{i\Delta \theta}
    \end{aligned}
\end{equation}
where $h^a,d^a$ are the height and diagonal length of the anchor box, respectively. {To simplify notation, we omit the superscript $t$ of $\hat{\sigma}$ and $\hat{\mu}$, which represents the scene.}

{The uncertainty $U^t(\cdot)$ for the entire scene $\mathcal{P}^{t}$ 
can be calculated using the following equation:}

\begin{equation}
    \label{eq:all_varance}
    \begin{aligned}
        U^t(\hat{\mathcal{M}}^t)=\frac{1}{7\hat{N}^t} \sum_{i=1}^{\hat{N}^t}\sum_\varphi  (\left[\hat{\sigma}_{i\varphi }^t\right]_{au}+\eta \left[ \hat{\sigma}^t_{i\varphi }\right]_{eu}),
    \end{aligned}
\end{equation}
where $\eta$ is the ratio of the two types of uncertainties, set to 0.5 in our experiments, and $\varphi \in \{x, y, z, w, h, l, \theta \}$.

\begin{algorithm}[h]
\caption{Three-Stage Joint Sampling Algorithm}
\label{alg:three_stage_alg}
\begin{algorithmic}[1]
\REQUIRE ~~\\ 
$\mathcal{D}_i$: Initial labeled dataset;\\
$\mathcal{D}_u$: Unlabeled dataset;\\
$\Omega$: The oracle that labels the data;\\
$f_m(\omega_r;\cdot)$: The predictor with MDN head;\\
$R$: Total number of the active learning rounds;\\
$N_r$: The number of searched scenes for each round;\\
$K_1,K_2$: The selection ratio in the first and second stages.
\ENSURE ~~\\
$f_m(\omega_R;\cdot)$: The trained predictor;\\
$\mathcal{D}_l$: The final labeled data.
\STATE $f_m(\omega_0;\cdot) \gets$ pretrained model on $\mathcal{D}_i$
\STATE $D_l \leftarrow D_i$
\FOR{each $r \in [1,R]$}
    \STATE $\mathcal{D}_{S_1} \leftarrow \emptyset$, $\mathcal{D}_{S_2} \leftarrow \emptyset$, $\mathcal{D}_{r} \leftarrow \emptyset$
    \STATE $\{(\hat{\mathcal{C}^t}, \hat{\mathcal{B}^t}, \hat{\mathcal{M}^t})\}_{t\in\{ 1,...,\left | \mathcal{D}_u \right | \}} \gets f_m(\omega_{r-1};\mathcal{D}_u)$ inference on $\mathcal{D}_u$ 
    \FOR{each $\mathcal{P}^t \in \mathcal{D}_u$}
        \STATE $[\mathcal{P}^t, \mathcal{E}^t]\leftarrow$ $E^t(\hat{\mathcal{C}}^t)$ \COMMENT{calculated by Eq. (\ref{eq:cate_entropy})}
    \ENDFOR
    \STATE $\mathcal{D}_{S_1} \gets$ sort $\mathcal{D}_u$ by $\mathcal{E}^t$ and select scenes on top $K_1N_r$
    \STATE $\mathcal{D}_{S_2}\leftarrow$ select $K_2N_r$ scenes by Alg. \ref{alg:farthest_sampling_alg} in $\mathcal{D}_{S_1}$
     \FOR{each $\mathcal{P}^t \in \mathcal{D}_{S_2}$}
        \STATE $[\mathcal{P}^t, \mathcal{U}^t] \leftarrow U^t(\hat{\mathcal{M}}^t)$ \COMMENT{obtained by Eq. (\ref{eq:all_varance})}
    \ENDFOR
    \STATE $\mathcal{D}_{S_3} \gets$ sort $\mathcal{D}_{S_2}$ by $\mathcal{U}^t$ and select scenes on top $N_r$
    \STATE $\mathcal{D}_{S_3} \gets \mathcal{D}_{r} $
    \STATE $\mathcal{D}^*_r \leftarrow \Omega(\mathcal{D}_{r})$ \COMMENT{label the scenes} 
    \STATE $\mathcal{D}_l \leftarrow \mathcal{D}_l \cup \mathcal{D}^*_{r}$
    \STATE $\mathcal{D}_u \leftarrow \mathcal{D}_u / \mathcal{D}_{r}$
    \STATE $f_m(\omega_r;\cdot)\leftarrow$ retrain the model on $\mathcal{D}_l$
\ENDFOR
\STATE \textbf{return} $f_m(\omega_R;\cdot)$ and $\mathcal{D}_{l}$
\end{algorithmic}
\end{algorithm}

\section{Multi-stage joint sampling Strategy}
\label{sec:selection}
Similar to \cite{luo2023exploring}, we adopt a joint selection strategy with three evaluation metrics in sequence. Unlike single-stage methods that directly combine metrics with weights, the multi-stage allocates distinct metrics to select high-quantity scenes stage by stage. 

The order in which metrics are utilized significantly impacts the final sampling outcome. Considering the time consumption of Algorithm \ref{alg:farthest_sampling_alg}, it's prudent to avoid placing the scene similarity in the initial stage when handling large-scale original data. For the other two metrics, we prioritize selecting category entropy in the first stage for two reasons: 1) from the perspective of data balance, initiating with uncertainty metrics may lead to a concentration of selected data near decision boundaries, potentially excluding many balanced samples and directly influencing the subsequent use of category entropy metrics; 2) accurate classification is fundamental for precise regression in the detection task. Given that the roles of category entropy and perceptual uncertainty metrics in this paper correspond to classification and regression, respectively, prioritizing category entropy metric first is reasonable logically.

Based on the preceding analysis, the selection process unfolds across multiple stages. Initially, we identify $K_1N_r$ scenes with the highest value based on the category entropy metric, forming $\mathcal{D}_{S_1}$ from the unlabeled dataset. Subsequently, in the second stage, we leverage the similarity metric and the furthest-point sampling in Algorithm \ref{alg:farthest_sampling_alg} to pick $K_2N_r$ scenes, as $\mathcal{D}_{S_2}$, from $\mathcal{D}_{S_1}$, thereby ensuring the diversity of the selected scenes. Moving to the third stage, we use the perception uncertainty metric to select $N_r$ scenes with the highest uncertainty, as $\mathcal{D}_{S_3}$, from $\mathcal{D}_{S_2}$. Finally, $N_r$ scenes are sampled as the culmination of the selection strategy, with the detailed steps shown in Algorithm \ref{alg:three_stage_alg}.

\begin{table*}[!t]
	\centering
	\caption{A performance comparison with other AL methods was conducted on the KITTI, Lyft, nuScenes and SUScape \textit{val} sets. By the end of the AL process, approximately 32\% of the KITTI dataset, 11\% of the Lyft dataset, 13\% of the nuScenes dataset, and 16\% of the SUScape dataset were annotated. We report $mAP_{3D}$ and $mAP_{BEV}$ across 40 recall positions for 3 classes in KITTI, 9 classes in Lyft,  10 classes in nuScenes, and 8 classes in SUScape. Note that due to the absence of occlusion data, the \textit{Hard difficulty level is marked as Not Available (-) for nuScenes and SUScape.}}
	\setlength{\tabcolsep}{3.5mm}{
		\begin{tabular}{l|c|ccc|ccc}
			\hline
			\multirow{2}{*}{Dataset}  & \multirow{2}{*}{Method}       & \multicolumn{3}{c|}{$mAP_{3D}$}    & \multicolumn{3}{c}{$mAP_{BEV}$}                                                                                                                                                         \\ \cline{3-8}
			                          &                               & Easy                               & Moderate                           & Hard                                & Easy                               & Moderate                           & Hard                               \\ \hline
			\multirow{7}{*}{KITTI}    & Random                        & \multicolumn{1}{c}{70.68}          & \multicolumn{1}{c}{58.75}          & \multicolumn{1}{c|}{55.28}          & \multicolumn{1}{c}{75.19}          & \multicolumn{1}{c}{64.40}          & \multicolumn{1}{c}{60.98}          \\
			                          & Confidence                    & \multicolumn{1}{c}{68.98}          & \multicolumn{1}{c}{57.17}          & \multicolumn{1}{c|}{53.40}          & \multicolumn{1}{c}{72.95}          & \multicolumn{1}{c}{62.95}          & \multicolumn{1}{c}{59.74}          \\
			                          & MC Dropout\cite{feng2019deep} & \multicolumn{1}{c}{71.98}          & \multicolumn{1}{c}{58.79}          & \multicolumn{1}{c|}{55.22}          & \multicolumn{1}{c}{76.68}          & \multicolumn{1}{c}{65.60}          & \multicolumn{1}{c}{61.94}          \\
			                          & Coreset\cite{sener2017active} & \multicolumn{1}{c}{70.57}          & \multicolumn{1}{c}{58.92}          & \multicolumn{1}{c|}{54.99}          & \multicolumn{1}{c}{75.09}          & \multicolumn{1}{c}{64.89}          & \multicolumn{1}{c}{61.65}          \\
			                          & Badge\cite{ash2019deep}       & \multicolumn{1}{c}{70.31}          & \multicolumn{1}{c}{58.81}          & \multicolumn{1}{c|}{54.78}          & \multicolumn{1}{c}{74.89}          & \multicolumn{1}{c}{64.55}          & \multicolumn{1}{c}{61.03}          \\
			                          & Crb\cite{luo2023exploring}    & \multicolumn{1}{c}{71.16}          & \multicolumn{1}{c}{59.29}          & \multicolumn{1}{c|}{55.72}          & \multicolumn{1}{c}{76.16}          & \multicolumn{1}{c}{65.61}          & \multicolumn{1}{c}{62.14}          \\
			                          & TSceneJAL(Ours)               & \multicolumn{1}{c}{\textbf{73.21}} & \multicolumn{1}{c}{\textbf{61.32}} & \multicolumn{1}{c|}{\textbf{57.54}} & \multicolumn{1}{c}{\textbf{77.51}} & \multicolumn{1}{c}{\textbf{66.91}} & \multicolumn{1}{c}{\textbf{63.23}} \\ \hline \hline
			\multirow{7}{*}{Lyft}     & Random                        & 30.62                              & 27.87                              & 27.27                               & 34.32                              & 32.58                              & 31.95                              \\
			                          & Confidence                    & 30.77                              & 27.31                              & 26.79                               & 34.42                              & 31.96                              & 31.51                              \\
			                          & MC Dropout\cite{feng2019deep} & 31.35                              & 28.18                              & 27.48                               & 35.44                              & 33.87                              & 32.84                              \\
			                          & Coreset\cite{sener2017active} & 30.96                              & 27.85                              & 27.17                               & 35.45                              & 33.41                              & 32.67                              \\
			                          & Badge\cite{ash2019deep}       & 29.92                              & 26.83                              & 26.63                               & 33.94                              & 32.06                              & 31.68                              \\
			                          & Crb\cite{luo2023exploring}    & 31.10                              & 27.88                              & 27.07                               & 35.57                              & 33.46                              & 32.43                              \\
			                          & TSceneJAL(Ours)               & \textbf{32.49}                     & \textbf{29.56}                     & \textbf{29.26}                      & \textbf{36.66}                     & \textbf{34.91}                     & \textbf{34.49}                     \\ \hline \hline
			\multirow{7}{*}{nuScenes} & Random                        & 12.59                              & 11.24                              & -                                   & 13.45                              & 12.02                              & -                                  \\
			                          & Confidence                    & 12.48                              & 11.13                              & -                                   & 13.37                              & 11.90                              & -                                  \\
			                          & MC Dropout\cite{feng2019deep} & 12.38                              & 11.17                              & -                                   & 13.35                              & 11.93                              & -                                  \\
			                          & Coreset\cite{sener2017active} & 12.92                              & 11.56                              & -                                   & 13.90                              & 12.39                              & -                                  \\
			                          & Badge\cite{ash2019deep}       & 13.42                              & 12.01                              & -                                   & 14.38                              & 12.84                              & -                                  \\
			                          & Crb\cite{luo2023exploring}    & 12.84                              & 11.27                              & -                                   & 13.76                              & 12.12                              & -                                  \\
			                          & TSceneJAL(Ours)               & \textbf{13.78}                     & \textbf{12.28}                     & -                                   & \textbf{14.90}                     & \textbf{13.29}                     & -                                  \\ \hline \hline
			\multirow{7}{*}{SUScape}  & Random                        & 37.86                              & 34.51                              & -                                   & 43.42                              & 41.63                              & -                                  \\
			                          & Confidence                    & 36.91                              & 33.22                              & -                                   & 42.41                              & 40.51                              & -                                  \\
			                          & MC Dropout\cite{feng2019deep} & 37.90                              & 34.60                              & -                                   & 43.48                              & 41.84                              & -                                  \\
			                          & Coreset\cite{sener2017active} & 38.51                              & 34.67                              & -                                   & 43.91                              & 42.56                              & -                                  \\
			                          & Badge\cite{ash2019deep}       & 39.17                              & \textbf{35.75}                     & -                                   & 43.89                              & 42.59                              & -                                  \\
			                          & Crb\cite{luo2023exploring}    & 38.09                              & 34.46                              & -                                   & 43.79                              & 42.38                              & -                                  \\
			                          & TSceneJAL(Ours)               & \textbf{39.31}                     & 35.62                              & -                                   & \textbf{44.50}                    & \textbf{43.02}                     & -                                  \\ \hline
		\end{tabular}}
	\label{al_result}
\end{table*}

\section{Experiments}
\label{sec:experiment}
\subsection{Experimental Setup}
\textbf{3D Object Detection Datasets}. 
In our experiments, we utilize the KITTI\cite{geiger2012we}, Lyft\cite{kes2019lyft}, nuScenes\cite{caesar2020nuscenes}, and SUScape\cite{suscape} datasets for 3D object detection. We adopt the same dataset split as described in \cite{openpcdet2020}. 
Specifically, the training sets contain 3,711, 12,016, 26,706, and 22,862 frames, while their training-to-validation ratios are 0.98, 4.15, 4.63, and 5.04 for KITTI, Lyft, nuScenes, and SUScape, respectively. 
We sample data from their respective training sets, and testing is performed on their corresponding validation sets. 
To standardize our experimentation, the Lyft, nuScenes, and SUScape datasets were converted to the KITTI format. The conversion for nuScenes and SUScape was done using officially provided scripts, while the Lyft dataset was converted following the same procedure outlined in \cite{wang2020train}.


\textbf{Evaluation Metrics}. We consistently apply the official KITTI metrics\cite{geiger2012we} to evaluate the performance of our perception model, enabled by the conversion of other datasets into KITTI format. For all experiments, we train on all categories specific to each dataset and report two types of mean average precision (mAP): $mAP_{BEV}$ based on bird’s-eye view (BEV) IoUs, and $mAP_{3D}$ based on 3D IoUs. 
It should be noted that, since the official conversion scripts for the nuScenes and SUScape datasets do not provide the \textbf{occlusion} and \textbf{truncation} attributes, which are used to categorize labels into \textit{moderate} and \textit{hard} difficulty levels in KITTI evaluation, the results for the \textit{hard} difficulty level for both nuScenes and SUScape are marked as N/A (not available).

\textbf{Implementation Details}. The predictor used in our experiments is PointPillar, implemented based on the OpenPCDet\cite{openpcdet2020} framework. When modifying the model's regression head, we opt for setting the number of Gaussian distributions to 3, balancing training efficiency and detection performance. For a fair comparison, all AL methods are based on this modified PointPillar architecture. 
During training, the batch sizes are set to 8, 4, 4, and 2 for the KITTI, Lyft, nuScenes, and SUScape datasets, respectively.
The Adam optimizer is selected, with the learning rate, weight decay, and momentum set to 0.003, 0.01, and 0.9, respectively. In the AL process, we first randomly select $N_0$ scenes for annotation and train the initial learner. Subsequently, we perform $R$ rounds of AL iterations, selecting $N_r$ scenes from unlabeled data in each round. When employing the joint sampling strategy in each round, we set $K_1=3, K_2=2.5$, ensuring the selection quantities for each stage as $3N_r, 2.5N_r, N_r$, respectively. 
The parameters for the datasets are as follows: for KITTI, $N_0 = 200$, $N_r = 200$, and $R = 5$; for Lyft, $N_0 = 400$, $N_r = 400$, and $R = 4$; for nuScenes, $N_0 = 600$, $N_r = 600$, and $R = 5$; and for SUScape, $N_0 = 600$, $N_r = 600$, and $R = 5$.
At the conclusion of the AL process, we annotated 32\% of the KITTI dataset, 11\% of Lyft, 13\% of nuScenes, and 16\% of SUScape, balancing computational cost with reliable model training. During the initial model training phase, the number of epochs is set to 50 for KITTI, 30 for SUScape and nuScenes, and 10 for Lyft. In each subsequent round, the number of epochs increases by 2.

\subsection{Main Results}
\subsubsection{\textbf{Baselines}}
We employed the modified PointPillars with MDN as the AL predictor and implemented common AL methods: (1) \textbf{Random}: a basic data sampling strategy, randomly selecting a certain number of samples from the unlabeled dataset in each AL round. (2) \textbf{Confidence}: typically used in classification tasks, selecting the samples with the lowest classification confidence. (3) \textbf{MC Dropout}\cite{feng2019deep}: employing the Monte Carlo method to estimate uncertainty in regression tasks. In our implementation, we activate dropout layers and calculate uncertainty by measuring the variance between multiple forward propagation results, selecting samples with the highest uncertainty. (4) \textbf{Coreset}\cite{sener2017active}: a diversity-focused AL method, constructing a core set of samples, in which a greedy algorithm is used to select samples with the maximum dissimilarity from the unlabeled dataset compared to the labeled data. (5) \textbf{Badge}\cite{ash2019deep}: a hybrid AL method that employs clustering techniques within the gradient embedding space to select samples. (6) \textbf{Crb}\cite{luo2023exploring}: a state-of-the-art AL method in 3D object detection, focusing on selection strategies of conciseness, representativeness, and balance.

\subsubsection{\textbf{Quantitative Analysis}}
The results of employing various AL methods are presented in Table \ref{al_result}. Our approach consistently outperforms all other methods in 3D object detection tasks across all datasets. Compared to \textit{Random}, our approach exhibits an average improvement of 2.45\% on KITTI, 2.13\% on Lyft, 1.24\% on nuScenes, and 1.25\% on SUScape within the same budget. When compared with the state-of-the-art method \textit{Crb}, our method achieves an improvement of 1.61\% on KITTI, 1.64\% on Lyft, 1.07\% on nuScenes, and 0.94\% on SUScape.

\begin{figure*}[!t]  
    \centering     
    \includegraphics[width=1.0\linewidth]{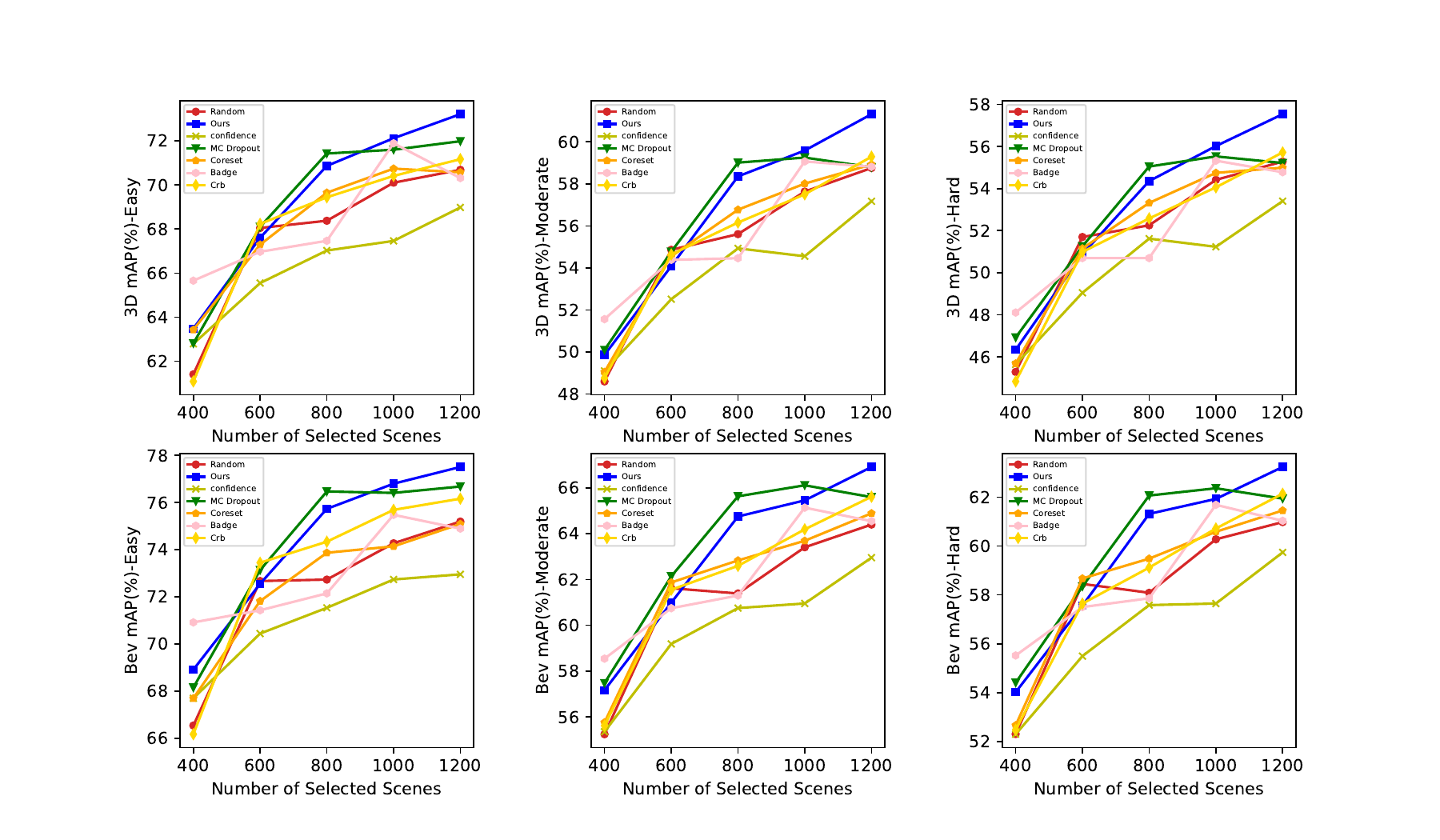}     
    \caption{Experiment results of different AL methods with an increasing number of sampled scenes on the KITTI \textit{val} set. Initially, all methods use the same set of 200 scenes. Over five iterations, 32\% of the whole data (1200 scenes) is selected. Notably, our method outperforms all other methods in the final two iterations in terms of $mAP_{3D}$.}   
    \label{fig:al_test_kitti} 
\end{figure*}
\begin{figure}[!h]  
    \centering     
    \includegraphics[width=1.05\linewidth]{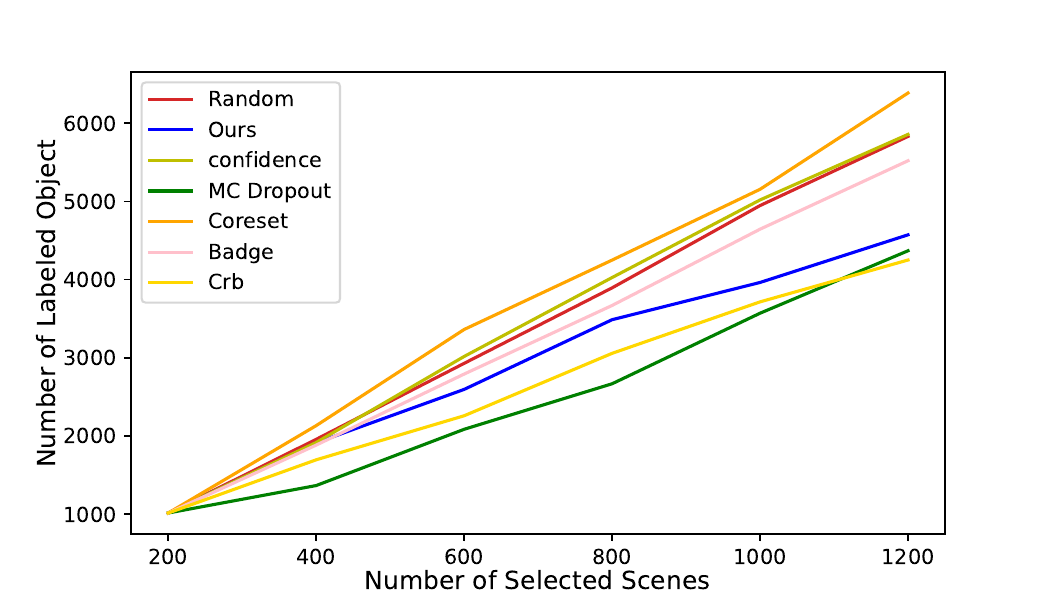}     
    \caption{Number of objects in the scenes selected by different AL methods on KITTI. It can be seen that, at the end of AL, the scenes selected by our method, \textit{MC Dropout}, and \textit{Crb} contain a relatively smaller number of objects, which will be accurately annotated by the oracle.}   
    \label{fig:al_test_kitti_number} 
\end{figure}

Under a fixed budget, the comparison results of different AL policies on 3D and BEV detection are shown in Fig. \ref{fig:al_test_kitti}. It can be seen that our method outperforms in the final two iterations in $AP_{3D}$, regardless of difficulty settings. Although our method does not yield optimal results in the first two iterations, it gradually selects data with great information as more frames join in, driving rapid performance improvements. Moreover, statistics on the number of selected bounding boxes are displayed in Fig. \ref{fig:al_test_kitti_number}. It can be observed that the annotation cost for our approach is approximately half that of \textit{Random} and \textit{Coreset} on the KITTI dataset, while achieving comparable performance. Besides, AL baselines for both regression and classification tasks (\textit{e.g., Ours, Crb,} and \textit{MC Dropout}) generally obtain higher performance but lower labelling costs, compared to the classification-only methods (\textit{e.g., Confidence}, and \textit{Badge}).

\subsubsection{\textbf{Qualitative Analysis}}
\begin{figure*}[!h]
    \centering     
    \includegraphics[width=1.0\linewidth]{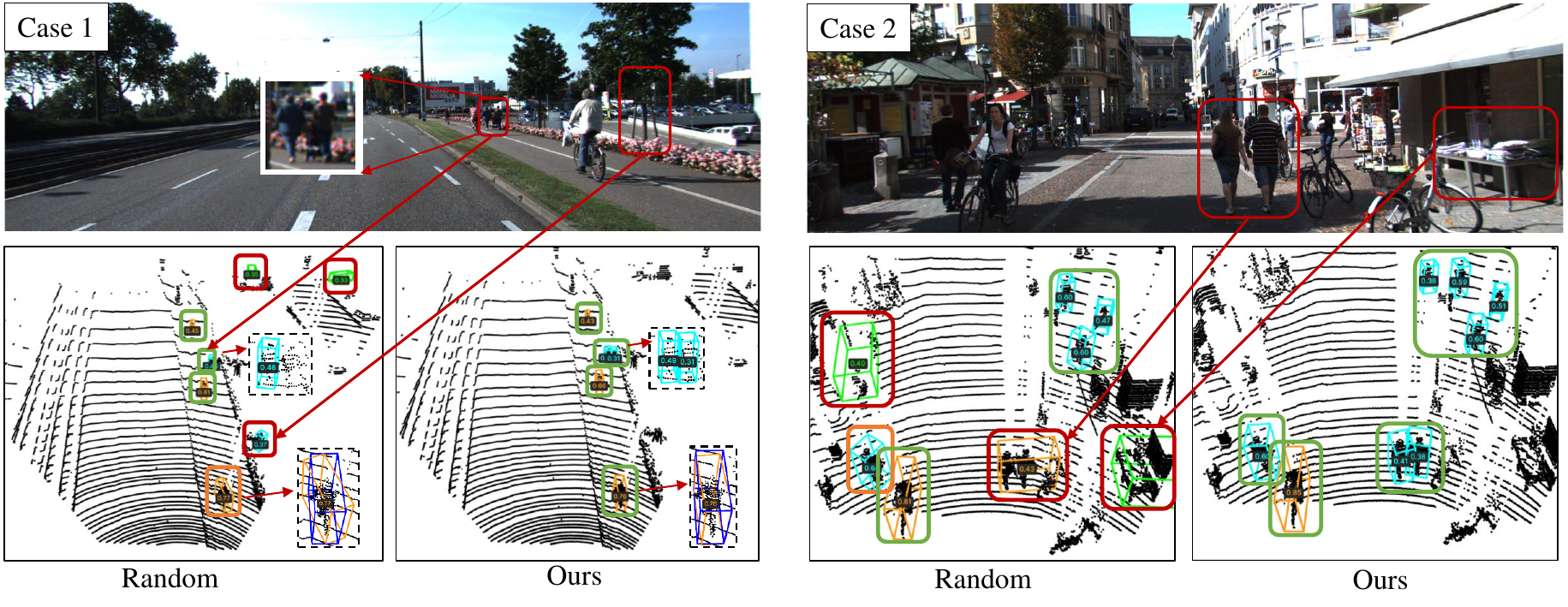}     
    \caption{Case study of 3D detection under \textit{Random} and \textit{TSceneJAL} frameworks, respectively. The detector is trained on approximately 32\% of the KITTI dataset. Green boxes represent true positives, red boxes represent false positives, and orange boxes indicate detections that are correct but have a lower IoU with the ground truth. It can be seen that our method produces a more accurate predictor for pedestrian and cyclist detection, which also demonstrates better resistance to noise interference, reducing false positives.}   
    \label{fig:case_study} 
\end{figure*}

Fig. \ref{fig:case_study} shows the detection outcomes of two cases predicted by our model and \textit{Random}. It can be seen that our AL approach can generate more accurate predictions with fewer false positives (indicated by red boxes), compared to \textit{Random}. For instance, in case 1, \textit{Random} mistakenly identifies a tree as a pedestrian, while in case 2, two walkers are incorrectly classified as cyclists by \textit{Random}. Besides, \textit{Random} achieves a significantly lower IoU scores with the ground truth compared to our method, resulting in predictions (indicated by orange boxes) with considerable orientation deviations.
These examples illustrate that under identical data conditions, our method trains a learner model with superior detection performance, particularly notable in classes with fewer instances (such as pedestrians and cyclists). Additionally, some instances of misclassification in the examples highlight the robustness of our method against noise interference as well. 
\label{sec:conclution}



\subsection{Ablation Study}

\begin{table}[]
\centering
\caption{Ablation study of different AL metrics on the KITTI \textit{val} set, where \textit{cate.}, \textit{simi.}, \textit{uncer.} stand for the category entropy, the scene similarity, and the perception uncertainty, respectively. We conducted experiments on the effects of three metrics and investigated the impacts of two types of uncertainties.}
\resizebox{\linewidth}{!}{
\begin{tabular}{cccccccccc}
\hline
\multirow{2}{*}{cate.} & \multirow{2}{*}{simi.} & \multicolumn{2}{c}{uncer.} & \multicolumn{3}{c}{$mAP_{3D}$} & \multicolumn{3}{c}{$mAP_{BEV}$} \\ \cline{3-10} 
                       &                        & AU           & EU          & Easy    & Mod.     & Hard   & Easy    & Mod.    & Hard    \\ \hline
-                      & -                      & -            & -           & 70.68   & 58.75   & 55.28  & 75.19   & 64.40   & 60.98   \\
-                      & -                      & $\checkmark$            & -           & 72.23   & 59.83   & 55.93  & 77.02   & 66.42   & 62.62   \\
-                      & -                      & -            & $\checkmark$           & 71.68   & 59.66   & 55.50  & 76.64   & 65.82   & 62.08   \\
-                      & -                      & $\checkmark$            & $\checkmark$           & 71.74   & 59.69   & 55.96  & 77.31   & 66.71   & 63.02   \\
$\checkmark$                      & -                      & -            & -           & 71.98   & 60.03   & 55.76  & 76.89   & 66.10   & 62.37   \\
$\checkmark$                      & $\checkmark$                      & -            & -           & 71.59         & 60.35         & 56.12        & 76.48         & 66.07         & 62.52         \\
$\checkmark$                      & $\checkmark$                      & $\checkmark$            & $\checkmark$           & \textbf{73.21} & \textbf{61.32} & \textbf{57.54} & \textbf{77.51} & \textbf{66.91} & \textbf{63.23}         \\ \hline
\end{tabular}}
\label{ablation_study}
\end{table}

\subsubsection{\textbf{Category Entropy Metric}}

\begin{figure}[!h]  
    \centering     
    \includegraphics[width=0.9\linewidth]{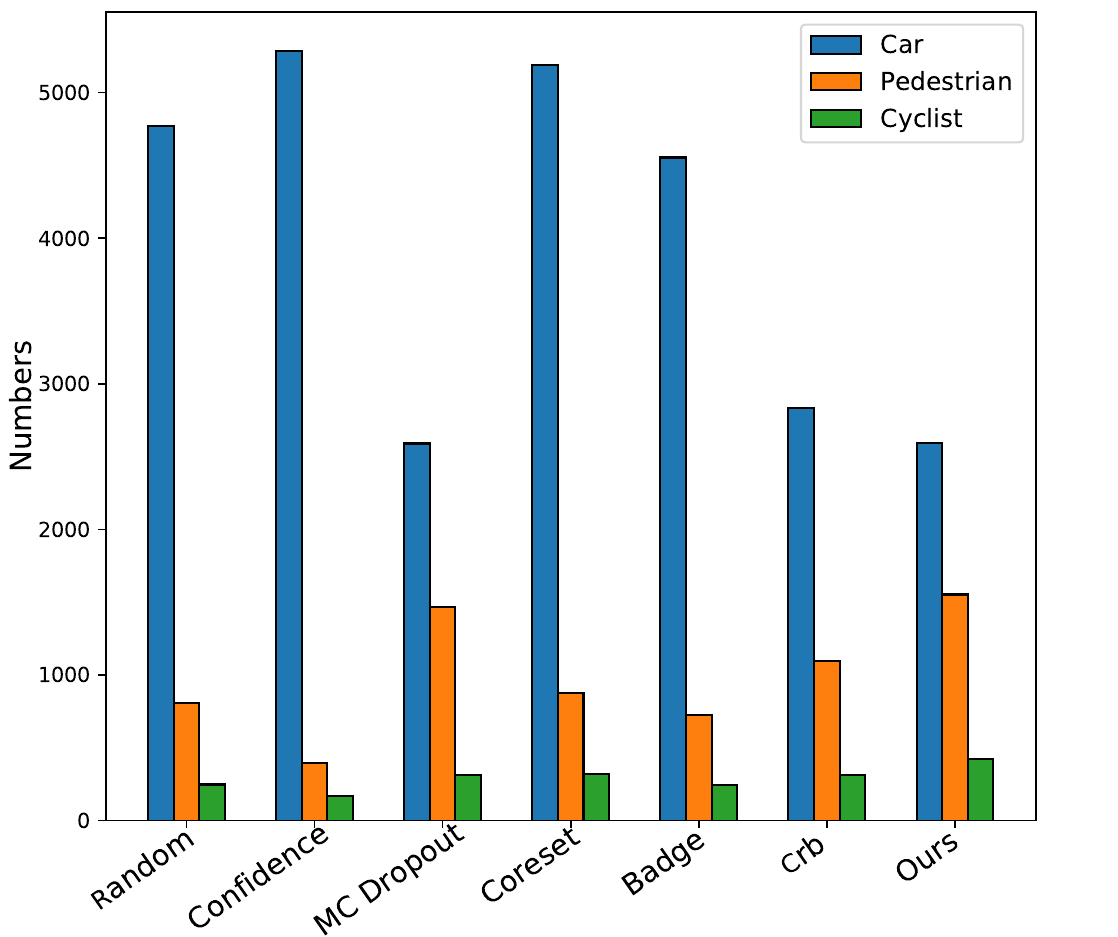}     
    \caption{Category distributions of the scenes selected by different AL methods on KITTI, including car, pedestrian, and cyclist. It can be seen that our method has superior ability to balance the distribution of categories. 
    }   
    \label{fig:kitti_cate_num} 
\end{figure}

In Table \ref{ablation_study}, it can be seen that using the category entropy alone as the AL sampler yields relatively promising results, with an average increase of +1.02 and +2.16 for 3D and BEV detection, respectively. Additionally, Table \ref{kitti_cate_result} presents performances of different methods concerning object classes on the KITTI dataset, in which our method is superior to others, but slightly lags behind \textit{Random} in terms of car category. The category entropy-based sampling scheme tends to choose scenes containing more classes, thereby enhancing the balance of different class quantities in the dataset. This trend is further demonstrated in Fig. \ref{fig:kitti_cate_num}, where our \textit{TSceneJAL} method can actively pick out more instances of pedestrian and cyclist. Hence, when designing AL selection policies, it is crucial to prioritize categories with fewer samples.

\begin{table*}[!t]
\centering
\caption{A comparison of detection results of three classes with different AL methods on the KITTI \textit{val} set. The table displays the $AP_{3D}$ over 40 recall positions for different categories at the end of all AL methods. The IoU thresholds for car, pedestrian, and cyclist are set to 0.7, 0.5, and 0.5, respectively.}
\setlength{\tabcolsep}{3.5mm}{
\begin{tabular}{cccccccccc}
\hline
                                               & \multicolumn{3}{c}{Car} & \multicolumn{3}{c}{Pedestrian} & \multicolumn{3}{c}{Cyclist} \\ \cline{2-10} 
Method                                         & Easy    & Mod.   & Hard  & Easy     & Mod.      & Hard     & Easy    & Mod.     & Hard    \\ \hline
Random                                         & 90.36 & \textbf{78.67} & \textbf{75.52} & 43.55    & 38.91    & 35.49    & 78.14   & 58.68   & 54.84   \\
Confidence                                     & 88.27   & 78.08 & 73.68 & 43.01    & 38.06    & 34.90    & 75.65   & 55.38   & 51.63   \\
MC Dropout\cite{feng2019deep}                                     & 89.92   & 75.77 & 72.76 & 50.08    & 45.10    & 40.99    & 75.93   & 55.50   & 51.90   \\
Coreset\cite{sener2017active} & \textbf{90.41}   & 78.54 & 74.21 & 43.47    & 39.47    & 35.90    & 77.83   & 58.75   & 54.87   \\
Badge\cite{ash2019deep}       & 88.42   & 78.19 & 73.86 & 45.71    & 40.58    & 36.84    & 76.79   & 57.67   & 53.62   \\
Crb\cite{luo2023exploring}    & 88.81   & 76.18 & 73.41 & 47.23    & 42.02    & 38.00    & 77.45   & 59.69   & 55.74   \\
TSceneJAL(Ours)                                           & 88.83   & 78.19 & 75.16 & \textbf{51.83} & \textbf{45.37} & \textbf{41.20} & \textbf{78.96} & \textbf{60.39} & \textbf{56.25}   \\ \hline
\end{tabular}}
\label{kitti_cate_result}
\end{table*}

\begin{figure}[!t]  
    \centering     
    \includegraphics[width=0.8\linewidth]{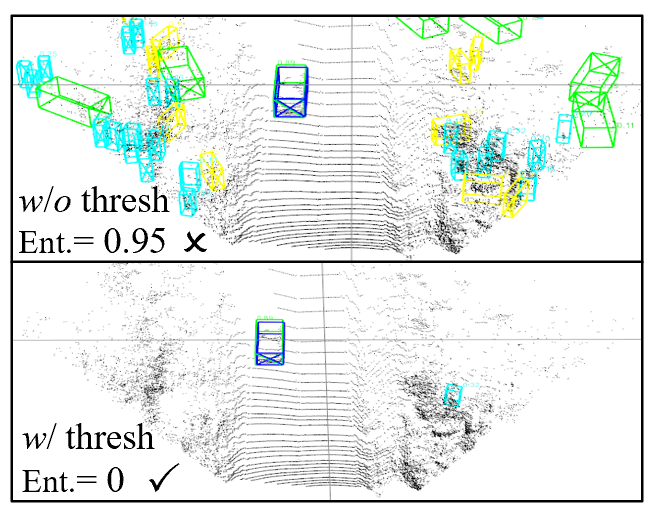}     
    \caption{An illustration of the impact of using a threshold during category entropy estimation. In a scene with only one car (blue box represents ground truth), the upper figure shows the prediction results without the threshold, revealing many incorrect detections, and a miscalculated category entropy of 0.95. The lower figure shows a correct prediction using the threshold (specified in Eq. (\ref{eq:delta_func})), with expected category entropy of 0.}   
    \label{fig:cate_thresh} 
\end{figure}

Category entropy is also applied in \textit{Crb}\cite{luo2023exploring}, but it neglects the fact that the AL predictor is hardly capable of detecting objects in the initial learning process. As shown in Fig. \ref{fig:cate_thresh}, due to various false positives predicted by the initial AL, the scene containing only one vehicle (with an expected entropy of 0) produces a high entropy value of 0.95, so that such data would be wrongly selected into the next stage with high probability. To address this issue, we filter out these scenes via a combination scheme of both class attribute and confidence threshold as Eq. (\ref{eq:delta_func}) and the gains of this filtering scheme are illustrated in Table \ref{tb:cate_thresh}.

\begin{table}[!t]
\centering
\caption{A comparison of detection performance, both with and without the utilization of the category confidence threshold. Solely using single-stage category entropy as the sampling strategy, \textbf{w/o} and \textbf{w/} indicate scenes where detection results are not filtered before category entropy calculation and are filtered with a threshold, respectively.}
\begin{tabular}{cccc}
\hline
           & \multicolumn{3}{c}{$mAP_{3D}$} \\ \cline{2-4} 
           & Easy   & Mod.    & Hard  \\ \hline
w/o thresh & 71.36  & 59.16  & 55.20  \\
w/ thresh  & 71.98(+0.62)  & 60.03(+0.87)  & 55.76(+0.56)  \\ \hline
\end{tabular}
\label{tb:cate_thresh}
\end{table}

\subsubsection{\textbf{Scene Similarity Metric}}

\begin{figure} [!t]
    \centering     
    \includegraphics[width=1.0\linewidth]{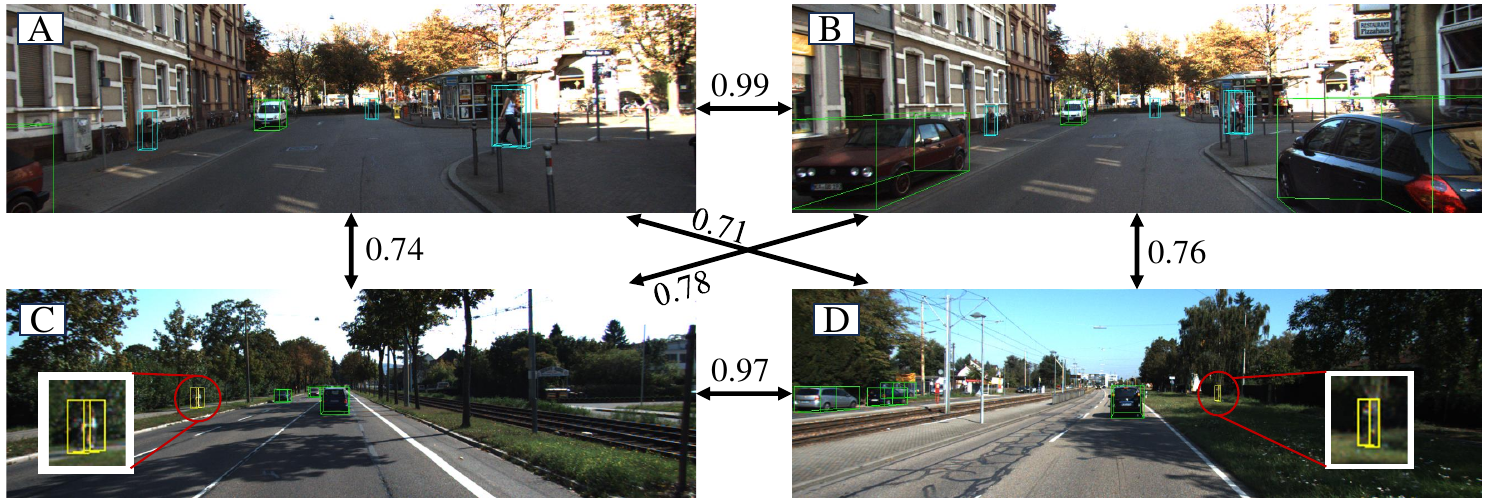}     
    \caption{An illustration of scene similarity from KITTI. Scenes A and B both contain cars (in \textcolor{green}{green}) and pedestrians (in \textcolor{blue}{blue}), exhibiting a high similarity. Scenes C and D, despite different object locations, both include cars and cyclists (in \textcolor{yellow}{yellow}) with relatively similar distributions. Conversely, scenes A and D, are different in categories, quantities, and distributions, resulting in the least similarity.}   
    \label{fig:simi_sample} 
\end{figure}

A similarity metric serves to quantify the distance between traffic scenes, and similarity-based sampling rests on the assumption that scenes with less similarity offer more diverse information for the active learner. As shown in Table \ref{ablation_study}, integrating the similarity sampling scheme after the entropy scheme yields a slight improvement. Furthermore, when uncertainty is combined, the performance is enhanced by about +0.9 $mAP_{3D}$ on average, compared to scenes without a similarity scheme. Besides, it can be seen from Table \ref{compare_simi} that a more significant improvement appears in the Lyft dataset, due to its sequential scenes introducing more redundant data than KITTI. The graph representation can effectively quantify the category quantity and distribution of traffic participants in each scene, providing a useful assessment for scene similarity at the frame level, some examples shown in Fig. \ref{fig:simi_sample}.  
\begin{table}[!t]
\centering
\caption{A comparison of the detection performance, applying a multi-stage sampling strategy across different datasets. Here, \textbf{w/o simi} denotes the utilization of a two-stage strategy without similarity evaluation; while \textbf{w/ simi} denotes the adoption of a three-stage strategy with similarity assessment.}
\begin{tabular}{ccccc}
\hline
                       &           & \multicolumn{3}{c}{$mAP_{3D}$}       \\ \cline{3-5} 
                       &           & Easy         & Mod.          & Hard         \\ \hline
\multirow{2}{*}{KITTI} & w/o simi.  & 72.77        & 60.54        & 56.05        \\
                       & w/ simi. & 73.21(+0.44) & 61.32(+0.78) & 57.54(+1.49) \\ \hline
\multirow{2}{*}{Lyft}  & w/o simi.  & 30.77        & 27.95        & 27.25        \\
                       & w/ simi. & 32.49(+1.72) & 29.56(+1.61) & 29.26(+2.01) \\ \hline
\end{tabular}
\label{compare_simi}
\end{table}

To facilitate the selection of more diverse scenes using similarity metrics, a greedy farthest sampling (FS) scheme is adopted, as illustrated in Algorithm \ref{alg:farthest_sampling_alg}. Table \ref{tab:fs_alg} presents a comparison between FS and random sampling during the similarity-based selection process. It is noted that the employment of the FS consistently achieves superior performance. Additionally, our algorithm conducts initial point sampling before sampling, which enhances algorithm stability. 

\begin{table}[!t]
\centering
\caption{A comparison of the detection performance between our farthest sampling (FS) scheme and random sampling scheme, following the first-stage category entropy sampling. \textbf{w/o init} denotes that no initial sampling point is specified in FS. Each experiment was conducted three times, and the reported accuracy includes the mean along with the standard deviation.}
\begin{tabular}{cccc}
\hline
           & \multicolumn{3}{c}{$mAP_{3D}$}       \\ \cline{2-4} 
           & Easy         & Mod.          & Hard         \\ \hline
 Random  & 62.03$\pm$0.98        & 48.63$\pm$0.88        & 45.10$\pm$\textbf{0.65}        \\
 FS w/o init & 63.08$\pm$1.47 & 50.04$\pm$1.40 & 46.44$\pm$1.25 \\ 
 FS  & \textbf{64.21}$\pm$\textbf{0.42}        & \textbf{50.66}$\pm$\textbf{0.75}        & \textbf{47.16}$\pm$0.66        \\ \hline
\end{tabular}
\label{tab:fs_alg}
\end{table}

\begin{figure}[!t]  
    \centering     
    \includegraphics[width=1\linewidth]{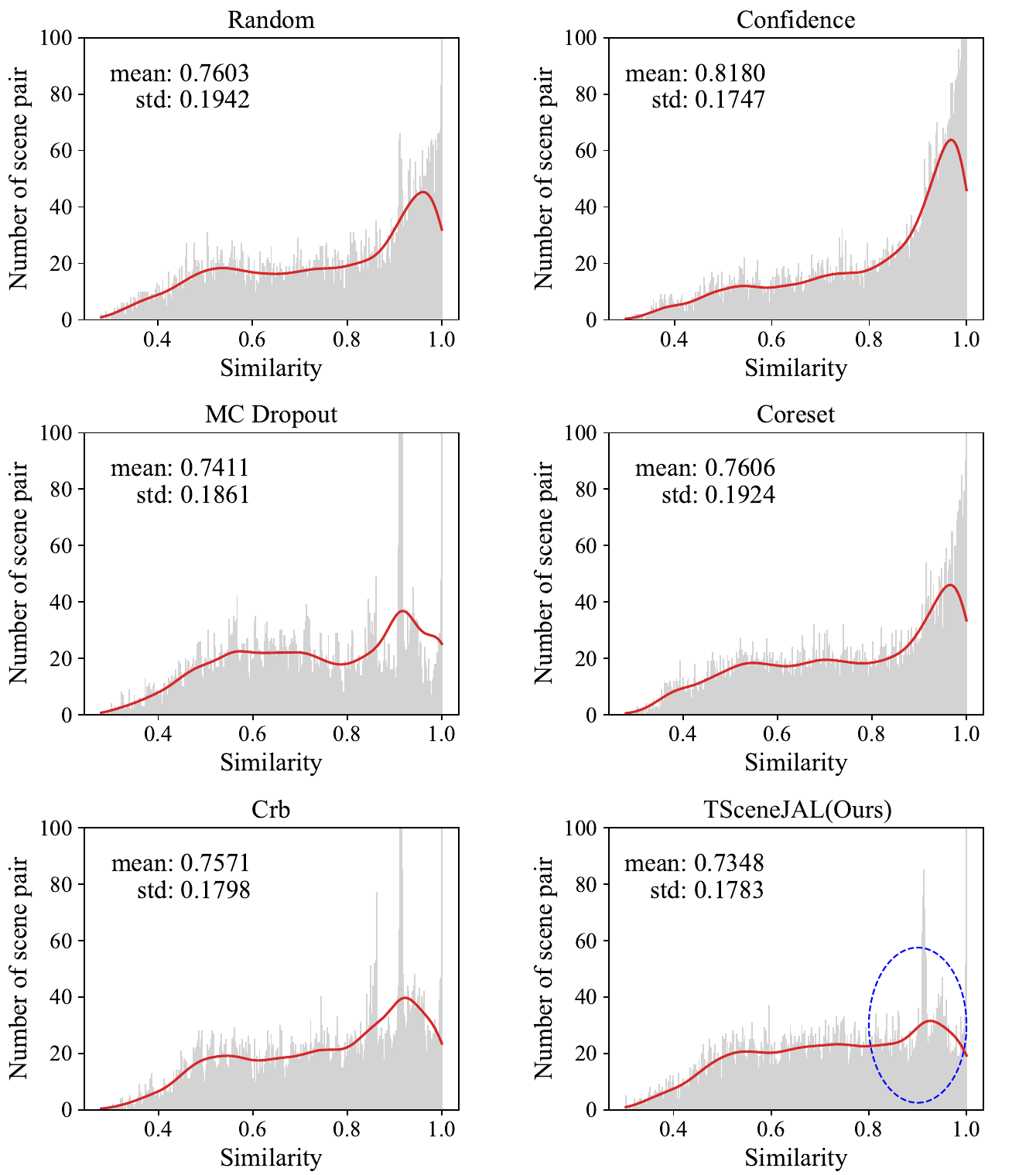}     
    \caption{Similarity distributions of the scenes selected by different AL
     methods on KITTI. We calculate the similarity of 10k scene pairs randomly extracted from the sampled candidates, with the $red$ line indicating the kernel density estimation of the corresponding distribution. It can be seen that the scenes chosen by our \textit{TSceneJAL} method display lower redundancy compared to others.}   
    \label{fig:simi_dis} 
\end{figure}

Furthermore, each AL method randomly extracts 10k paired scenes from the labeled pool, and their scene similarities are distributed in Fig. \ref{fig:simi_dis}. Our \textit{TSceneJAL} method demonstrates the ability to incorporate various similar scenes in a balanced manner, with relatively low mean and standard deviation of scene similarity, thus preventing the accumulation of highly similar scenes.

\subsubsection{\textbf{Perceptual Uncertainty Metric}}
\begin{figure}[!t]  
    \centering     
    \includegraphics[width=1.0\linewidth]{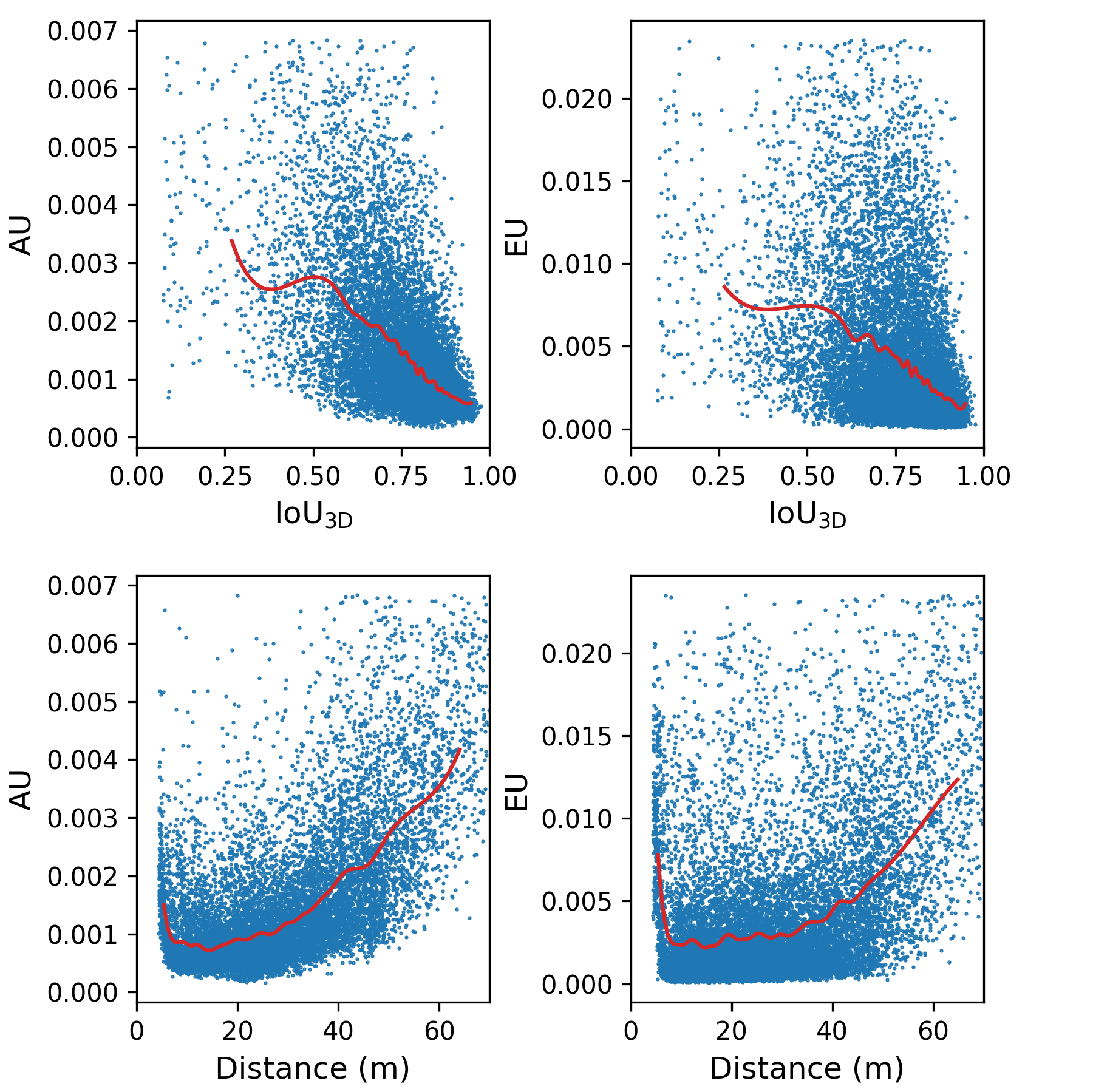}     
    \caption{An illustration of the changes of perception uncertainty in terms of IoU and sensor-to-object distance. The vertical axis is either aleatoric uncertainty (AU) or epistemic uncertainty (EU). The \textit{red} line in the scatter plot denotes the mean uncertainty, offering insights into the correlation between the two variables. Notably, both uncertainties are positively correlated with distance ($\ge$10m) and negatively correlated with IoU, consistent with our anticipated outcomes.}   
    \label{fig:uncer_1} 
\end{figure}

\begin{table}[!t]
\centering
    \caption{Comparison of the average AU and EU for scenes selected by different AL methods on KITTI, where \textbf{Avg.} denotes the average uncertainty per scene, with numerical values presented in $10^{-2}$ order. It can be seen that our method stands out for selecting scenes with the highest AU and EU.}
    \label{tab:uncer_select_stat}
\resizebox{\linewidth}{!}{\begin{tabular}{cccccccc}
\hline
& Random & Conf. & \thead{MC \\Dropout} & Coreset & Badge & Crb   & \thead{TSceneJAL\\(ours)}      \\ \hline
Avg. AU & 0.663  & 0.474      & 0.842       & 0.646   & 0.678 & 0.851 & \textbf{0.954} \\
Avg. EU & 1.732  & 1.156      & 2.594       & 1.806   & 1.772 & 1.887 & \textbf{2.608} \\ \hline
\end{tabular}}
\end{table}

Choosing scenes with higher AU or EU can bring more valuable information for the active learner. As shown in Table \ref{ablation_study}, It is worth noting that focusing solely on AU yields a more substantial improvement in the final performance compared to EU, as AU tends to fluctuate with noise (\textit{e.g.}, object occlusions and sparse points) inherent in diverse dataset. Moreover, the combination policy of AU and EU via a hyper-parameter achieves superior performance. 
As our approach harnesses the capabilities of the MDN, both uncertainties can be obtained simultaneously in a single forward propagation, which confers an efficiency advantage over methods that need multiple sampling rounds, such as MC Dropout. Integrating the uncertainty metric into the entire sampling framework markedly enhances detection accuracy on both 3D and BEV, as shown in Table \ref{ablation_study}. This indicates that even after the two-stage selection process of entropy and similarity, some scenes still lack sufficient information, thereby increasing annotation costs without commensurately enhancing the learner's detection ability. By incorporating uncertainty metrics in the third stage, we can effectively mitigate the impact of these less informative scenes.

To delve deeper into the analysis of the AU and EU derived from the MDN, we illustrate the relationships with the sensor-to-object distance, as well as IoU, as shown in Fig. \ref{fig:uncer_1}. The scene uncertainties are presented as scatter points, with their means depicted as a red line. It can be seen that the uncertainty reveals a negative correlation in terms of IoU, that is, more accurate prediction is generally along with lower uncertainty. Meanwhile, the correlation with distance initially declines, followed by a notable increase, since objects nearer to LiDAR sensor suffer from truncation while those farther away possess sparser point clouds. The variations in these two uncertainties align closely with theoretical expectations, demonstrating the effectiveness of our uncertainty quantification.

Additionally, we compute the average uncertainty of scenes selected by different AL methods, as shown in Table \ref{tab:uncer_select_stat}. For a fair comparison, all uncertainties are obtained by the MDN model trained on the complete KITTI dataset. It can be seen that our \textit{TSceneJAL} can select scenes with higher AU and EU, indicative of their higher complexity compared to those chosen by other methods.

\begin{table}[!t]
\centering
\caption{Comparison of detection performance with various three-stage sequences, where $E$ denotes category entropy Metric, $S$ denotes scene similarity metric, and $U$ denotes perceptual uncertainty metric.}
\label{tab:stage_order}
\begin{tabular}{ccccccccc}
\hline
\multirow{2}{*}{$S1$}    &\multirow{2}{*}{$S2$}    &\multirow{2}{*}{$S3$}   & \multicolumn{3}{c}{$mAP_{3D}$}       & \multicolumn{3}{c}{$mAP_{BEV}$} \\ \cline{4-9} 
    &    &       & Easy         & Mod.          & Hard        & Easy         & Mod.         & Hard \\ \hline
 $E$&  $S$  &  $U$ & \textbf{73.21} & \textbf{61.32} & \textbf{57.54} & 77.51 & 66.91 & 63.23        \\
 $E$&  $U$  &  $S$ & 73.13 & 61.12 & 57.10  & 77.07  & \textbf{67.23}  & \textbf{63.42} \\ 
 $U$&  $E$  &  $S$ & 72.54 & 60.71 & 56.55  & \textbf{77.75}  & 66.84  & 63.14  \\ 
 $U$&  $S$  &  $E$ & 71.77 & 59.86 & 56.24  & 76.84  & 66.71  & 63.03 \\ \hline
\end{tabular}
\end{table}

\subsection{Analysis on the Sequence of 3 Sampling Stages}
In this paper, we implement a progressive reduction scheme to instance selection by sequentially applying different metrics during sampling strategies. The order of these evaluation metrics plays a pivotal role in the effectiveness of sampling. Table \ref{tab:stage_order} demonstrates how different orders of metrics impact the final results. Prioritizing category entropy metrics in the first stage yields better results compared to perceptual uncertainty, consistent with our insights into the multi-stage sampling method. Regarding the sequence of scene similarity metrics and perceptual uncertainty metrics in the second and third stages, the experimental results from the first two rows of Table \ref{tab:stage_order} indicate that minimal disparity in optimizing the final dataset. However, considering the more important $mAP_{3D}$ metric in 3D object detection, placing scene similarity metrics first is preferable.

\subsection{Analysis of Various Initial Data Quantity}
\begin{figure} 
    \centering
    \includegraphics[width=\columnwidth]{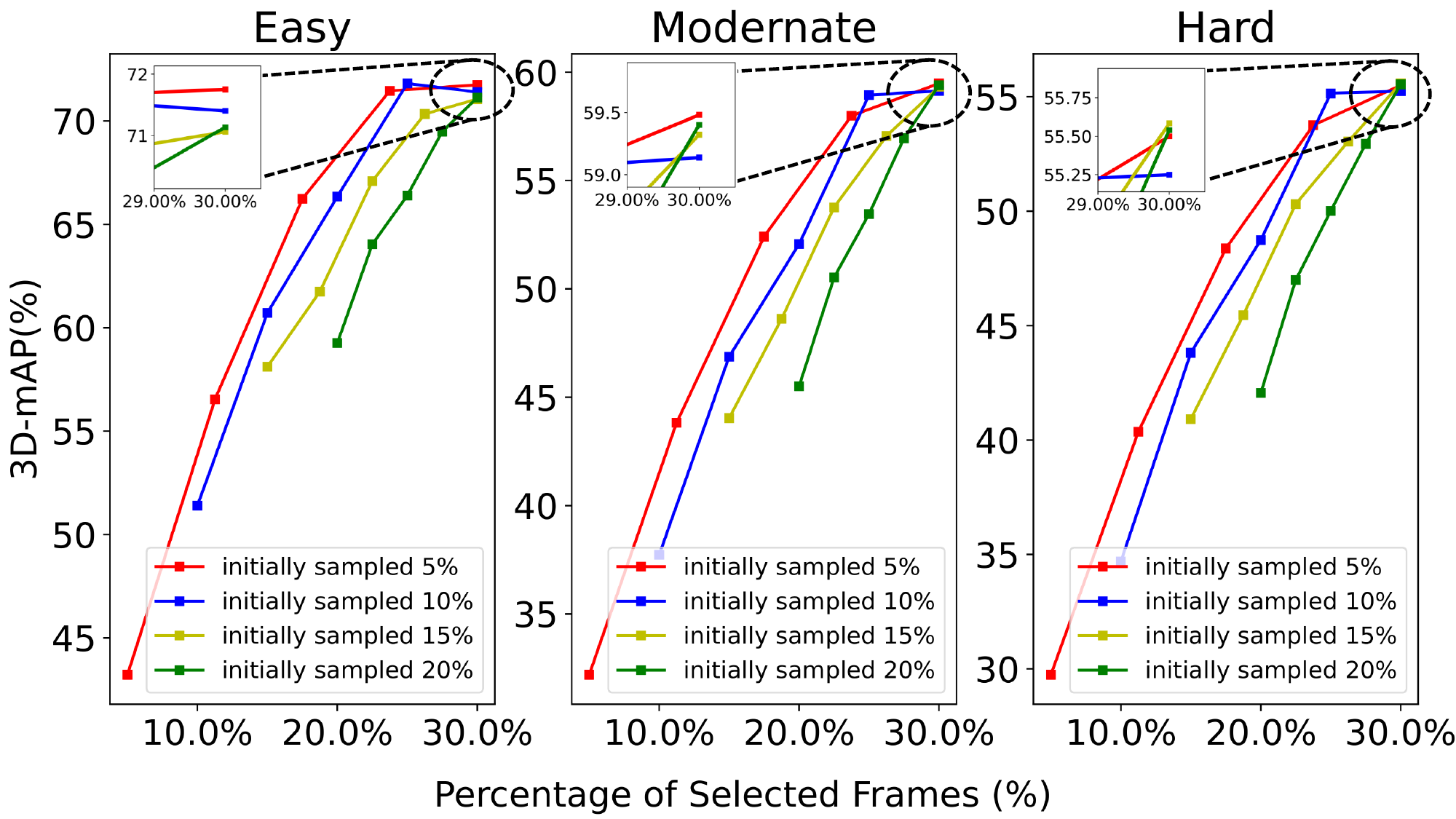}
    \caption{Experiment results of various initial random sampling size. To ensure fairness, all experiments ultimately sample 30\% of the total frames by adjusting the initial sampling size and the subsequent sampling size accordingly.}
    \label{fig:al_test_initsize}
\end{figure}

The initial model is trained with some random sampling data to get a foundational detection capability. Here, we further analyze the impact of the initial data volume on the model’s final performance. The initial data is randomly selected from 5\%, 10\%, 15\%, 20\% of the KITTI dataset for training different initial models, and after our \textit{TSceneJAL} sampling, the final number of selected samples is fixed to the same, 30\% of the KITTI, to ensure consistency. The results in terms of $mAP_{3D}$ are presented in Fig. \ref{fig:al_test_initsize}. It can be seen that although the initial models are sensitive to different data quantities in the first-stage selection, after multiple rounds of sampling, they can ultimately achieve similar performances, demonstrating the robustness of our \textit{TSceneJAL}. Besides, the larger the initial amount of data, the more training iterations these data are fed into the model, which is beneficial for fully training the data in handling some challenging scenarios, such as the models with 15\% and 20\% initial data achieving higher accuracy for ``Hard'' in Fig. \ref{fig:al_test_initsize}. However, this leads to a decreased space of expanding the valuable new data. After all, the number of training iterations can be increased as needed.

\begin{table}[]
{\centering
\caption{A comparison of selection algorithm time complexity and measured time costs for different strategies. \textit{Inference} show if this strategy need model inference in remained unlabeled dataset. \textit{Running Time} was evaluated on the KITTI dataset during the first active sampling process, conducted on a 64-core server equipped with an NVIDIA L40 GPU.}
\label{tab:time_cost}
\resizebox{\columnwidth}{!}{%
\begin{tabular}{cccc}
\toprule
Strategy   & Inference & Selection Algorithm Complexity                     & Running Time (s) \\
\midrule
Random     &          & $\mathcal{O}(N_r)$                            & 1.3               \\
Confidence & \checkmark         & $\mathcal{O}(n\log n)$                        & 183.9             \\
MC Dropout & \checkmark         & $\mathcal{O}(n\log n)$                        & 157.1             \\
Coreset    & \checkmark         & $\mathcal{O}(N_r n)$                          & 533.9             \\
Badge      & \checkmark         & $\mathcal{O}(N_r n)$                          & 427.4             \\
Crb        & \checkmark         & $\mathcal{O}(n\log n + 2N_r^2)$               & 256.7             \\
Ours       & \checkmark         & $\mathcal{O}(n\log n + N_r^2 + 2N_r \log N_r)$ & 216.6             \\
\bottomrule
\end{tabular}%
}\par}
\vspace{1mm}
$n$: Number of unlabaled frames.
\end{table}

\subsection{Algorithm Complexity Analysis}
Table \ref{tab:time_cost} presents the algorithm complexity and running time of various AL methods at the active selection stage. Specifically, \textit{Random} involves only the random generation of $Nr$ indices to retrieve samples from the pool. \textit{Confidence} and \textit{MC Dropout} sort the model-predicted scores, being $\mathcal{O}(n \log n)$. \textit{Coreset} utilizes pairwise distances between selected and unlabeled samples, with a time complexity of  $\mathcal{O}(N_rn)$, while similarly \textit{Badge} employs K-Means clustering for selection. \textit{Crb} uses multi-stage sampling within a single iteration, leading to a mixed time complexity of $\mathcal{O}(n \log n + 2N_r^2)$\cite{luo2023exploring}. Our \textit{TSceneJAL} approach also uses a multi-stage sampling process, and its time complexity includes three parts: (1) selecting $K_1N_r$ scenes from $n$ based on entropy sorting, $\mathcal{O}(n \log n)$; (2) selecting $K_2N_r$ scenes from $K_1N_r$ through pairwise similarity calculation and sorting, with complexity $\mathcal{O}(N_r^2 + N_r \log N_r)$; and (3) selecting $N_r$ scenes from $K_2N_r$ based on perception uncertainty, with complexity $\mathcal{O}(N_r \log N_r)$. Notably, in most AL selections, where $N_r \ll n$, the complexity $\mathcal{O}(n\log n + N_r^2 + 2N_r \log N_r)$ remains lower than $\mathcal{O}(N_r n)$. Consequently, our \textit{TSceneJAL} achieves efficient selection with state-of-the-art performance. Furthermore, the running times of these selection algorithms tested on the KITTI dataset are recorded in Table \ref{tab:time_cost}.

\subsection{Further Experiment on More Scenes}
\begin{figure}[!t]  
    \centering     
    \includegraphics[width=1.0\linewidth]{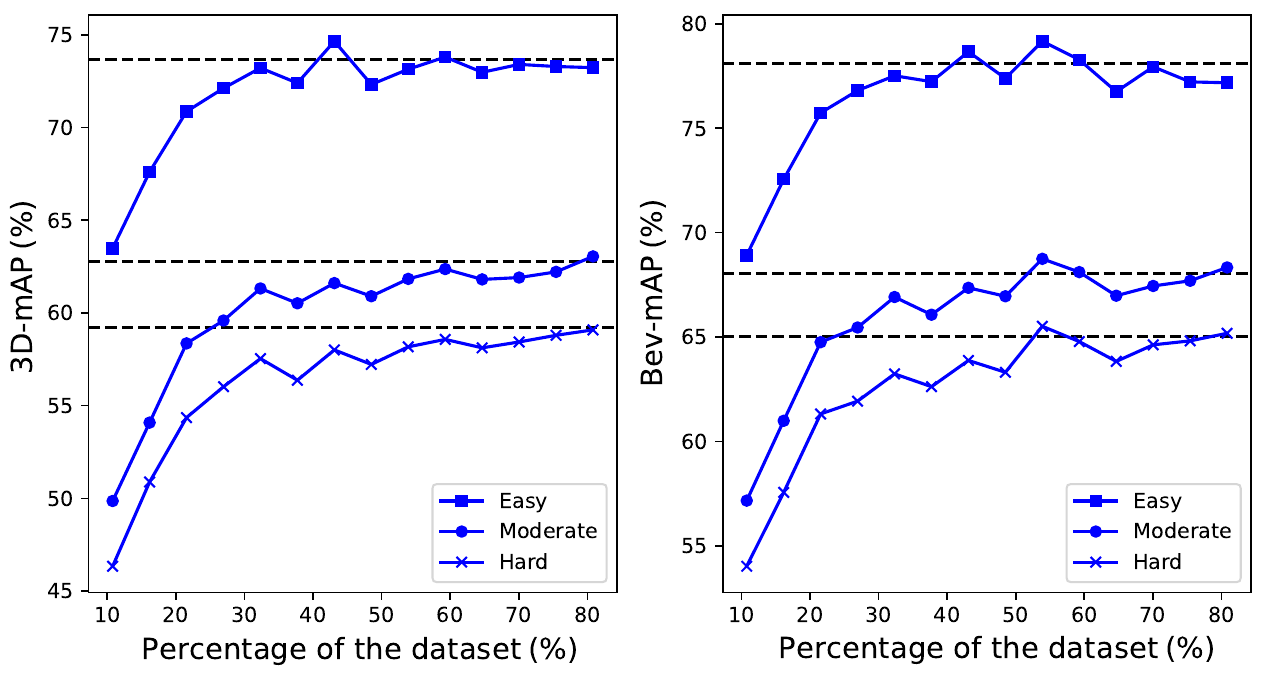}     
    \caption{Comparison of the performances of our \textit{TSceneJAL} and fully supervised learning (FSL) on the KITTI validation set. Dashed lines represent FSL results under the different difficulty levels: easy, moderate, and hard. Notably, as the dataset size reaches 80\%, the detection results in terms of 3D and Bev mAP are already comparable to or even surpass the results obtained with FSL.} 
    \label{fig:al_result_full} 
\end{figure}

To explore the impact of incorporating more scenes in active learning, we extend the AL process using the same experimental setup on KITTI. Fig. \ref{fig:al_result_full} illustrates the trend of detection accuracy under different difficulty levels as data volumes grow, compared to the upper bounds of fully supervised learning (FSL, dashed lines in Fig. \ref{fig:al_result_full}). It can be seen that the performances increase gradually after 30\%, eventually surpassing the upper bounds when the amount of data is between 60\% and 80\%, either on 3D or BEV detection. This means that 20\%$\sim$40\% of the KITTI dataset may be redundant, enabling savings in annotation costs. Furthermore, if only considering "Easy" scenes, even more redundant data can be removed.

\subsection{Further Experiment on Confidence Thresholds}
Using different confidence thresholds for sampling experiments, the results are presented in Table \ref{tab:thresh_exp2}. It is observed that the sampling performance is optimal at 0.3. Interestingly, setting the threshold to 0.9 results in poorer performance compared to having no threshold. This is due to excessively filtering out targets at the high threshold, causing almost identical results for each frame during entropy calculation, effectively resembling random sampling. The results show that at a threshold of 0.9, the outcome closely resembles the random sampling results shown in Table \ref{ablation_study}.

\begin{table}[!t]
\centering
\caption{A comparison of detection performance, with difference category confidence threshold. Solely using single-stage category entropy as the sampling strategy.}
\label{tab:thresh_exp2}
\begin{tabular}{clcll}
\hline
\multicolumn{1}{l}{}       & \multicolumn{1}{c}{}     & \multicolumn{3}{c}{$mAP_{3D}$}                                                    \\ \cline{3-5} 
\multicolumn{2}{l}{}                                  & Easy                      & \multicolumn{1}{c}{Mod.}  & \multicolumn{1}{c}{Hard}  \\ \hline
\multicolumn{2}{c}{w/o thresh}                        & 71.36                     & \multicolumn{1}{c}{59.16} & \multicolumn{1}{c}{55.20} \\ \hline
\multirow{4}{*}{w/ thresh} & \multicolumn{1}{c|}{0.3} & 71.98                     & \multicolumn{1}{c}{60.03} & \multicolumn{1}{c}{55.76} \\
                           & \multicolumn{1}{l|}{0.5} & \multicolumn{1}{l}{71.21} & 59.21                     & 55.75                     \\
                           & \multicolumn{1}{l|}{0.7} & \multicolumn{1}{l}{71.29} & 59.47                     & 55.78                     \\
                           & \multicolumn{1}{l|}{0.9} & \multicolumn{1}{l}{70.68} & 58.48                     & 55.03                     \\ \hline
\end{tabular}
\end{table}

\section{Conclusion}
\label{sec:conclution}


This paper has presented a joint active learning framework to select most informative scenes in an unlabeled dataset (or a newly collected dataset) for 3D object detection through a three-stage sampling scheme. The proposed sampling scheme includes three metrics: category entropy, scene similarity, and perception uncertainty. The proposed category entropy metric can help balance the distribution of different categories of traffic scenes. The proposed scene similarity metric can enrich the diversity of traffic scenes by using the directed graph scene representation. The proposed perception uncertainty metric can help capture the most complex traffic scenes by using a mixture density network (MDN) to compute both aleatoric and epistemic uncertainties. Experimental results show that our approach outperforms existing state-of-the-art methods with PointPillars on all four datasets, achieving average improvements of 2.5\%$\sim$6.6\%, 4.4\%$\sim$9.0\%, 3.0\%$\sim$10.9\% and 0.5\%$\sim$6.2\% in terms of $mAP_{3D}/mAP_{BEV}$  for KITTI, Lyft, nuScenes, and SUScape, respectively. Our future work includes extending the proposed AL framework to more downstream tasks, such as trajectory prediction and decision-making.


\section{Acknowledgements}
This work is jointly supported by 
the Hetao Shenzhen-HongKong Science and Technology Innovation Cooperation Zone and the Shenzhen Deeproute.ai Co., Ltd (HZQB-KCZYZ-2021055), 
the National Natural Science Foundation of China (62261160654),
the Shenzhen Fundamental Research Program (JCYJ20220818103006012, KJZD20231023092600001),
the Shenzhen Key Laboratory of Robotics and Computer Vision (ZDSYS20220330160557001)
and Science and Technology Development Fund (FDCT) Grants 0123/2022/AFJ and 0081/2022/A2.

\bibliographystyle{IEEEtran}
\bibliography{references}

\end{document}